\documentclass[preprint,12pt]{elsarticle}

\usepackage{lineno}
\usepackage{graphicx}
\usepackage{amsmath,amssymb}
\usepackage{booktabs,longtable,array}
\usepackage{hyperref}
\usepackage{xurl}
\usepackage[utf8]{inputenc}
\usepackage[T1]{fontenc}
\usepackage{textcomp}

\makeatletter
\providecommand\UseTaggingSocket[1]{}
\providecommand\@expl@@@mark@update@singlecol@structures@@{}
\def\LT@output{%
  \ifnum\outputpenalty <-\@Mi
    \ifnum\outputpenalty > -\LT@end@pen
      \LT@err{floats~ and~ marginpars~ not~ allowed~ in~ a~ longtable}\@ehc
    \else
      \setbox\z@\vbox{\unvbox\@cclv}%
      \ifdim \ht\LT@lastfoot>\ht\LT@foot
        \dimen@\pagegoal
        \advance\dimen@\ht\LT@foot
        \advance\dimen@-\ht\LT@lastfoot
        \ifdim\dimen@<\ht\z@
          \setbox\@cclv\vbox{\unvbox\z@\copy\LT@foot\vskip 0pt plus \maxdimen minus \normalbaselineskip}%
          \@makecol
          \@expl@@@mark@update@singlecol@structures@@
          \@outputpage
          \global\vsize\@colroom
          \setbox\z@\vbox{\box\LT@head}%
        \fi
      \fi
      \unvbox\z@\box\ifvoid\LT@lastfoot\LT@foot\else\LT@lastfoot\fi
      \UseTaggingSocket{tbl/longtable/foot}%
    \fi
  \else
    \setbox\@cclv\vbox{\unvbox\@cclv\copy\LT@foot\vskip 0pt plus \maxdimen minus \normalbaselineskip}%
    \UseTaggingSocket{tbl/longtable/foot}%
    \@makecol
    \@expl@@@mark@update@singlecol@structures@@
    \@outputpage
    \global\vsize\@colroom
    \copy\LT@head\nobreak
  \fi}
\makeatother

\journal{Computer Speech \& Language}
\hypersetup{pageanchor=false}
\makeatletter\def\ps@pprintTitle{\let\@oddhead\@empty\let\@evenhead\@empty\def\@oddfoot{\footnotesize\itshape Accepted manuscript. Published in Computer Speech \& Language 104 (2027) 102076.\hfill}\let\@evenfoot\@oddfoot}\makeatother
\begin{document}

\begin{frontmatter}

\title{Phonological Subspace Collapse Is Aetiology-Associated at the Group Level and Shape-Stable Across Languages: Evidence from 3,374 Speakers}

\author[smf]{Bernard Muller}
\author[tec]{Antonio Armando Ortiz Barrañón}
\author[smf]{LaVonne Roberts\corref{cor1}}
\ead{lavonne@scottmorganfoundation.org}
\cortext[cor1]{Corresponding author.}

\affiliation[smf]{organization={The Scott-Morgan Foundation},
                  city={Torquay},
                  country={United Kingdom}}
\affiliation[tec]{organization={Tecnológico de Monterrey},
                  city={Monterrey},
                  country={Mexico}}

\cortext[cor2]{Declaration of competing interest: The authors declare that they have no known competing financial interests or personal relationships that could have appeared to influence the work reported in this paper.}
\cortext[cor3]{Funding: This work was funded by The Scott-Morgan Foundation.}

\begin{abstract}
\noindent\textit{This is the accepted manuscript of the article published in Computer Speech \& Language 104 (2027) 102076, \url{https://doi.org/10.1016/j.csl.2026.102076}. \copyright\ 2026. This manuscript version is made available under the CC-BY-NC-ND 4.0 license \url{https://creativecommons.org/licenses/by-nc-nd/4.0/}.}

\medskip

We extend our training-free method for measuring phonological feature separability (d-prime) in frozen self-supervised speech representations to 3,374 speakers from 25 datasets in 12 languages, using six backbones. The main comparisons cover Parkinson's disease (PD), cerebral palsy (CP), amyotrophic lateral sclerosis (ALS), Down syndrome (DS), stroke and healthy controls. First, across these six groups, 10 of 13 features show large effects, and PD differs from CP, DS, and stroke on composite consonant d-prime. Severity mix may contribute to this separation. Within recorded severity categories, feature differences persist among mild speakers, including CP, ALS, and DS speakers within one English corpus. Pooled moderate speakers show little consonant separation. The apparent PD profile is sensitive to the healthy-control reference and remains confounded with language and corpus. Second, cross-language profile similarity is high for PD, CP and ALS (mean cosine 0.978--0.987), but random directions also produce high cosines (about 0.93). A prespecified test of aetiology-specific cross-language stability was inconclusive. Absolute d-prime values are not calibrated across languages. Third, composite results agree across backbones (inter-model rho = 0.83--0.99), while fine-grained feature orderings replicate only partially. Using reconstructed healthy-control directions and the main estimator, fixed-token estimates averaged over 50 draws retain severity correlations of -0.762 to -0.754 among the 467 speakers qualifying at every budget. Individual-level classification remains limited. These findings support population-level comparisons among the represented clinical groups, not individual screening.
\end{abstract}


\end{frontmatter}
\hypersetup{pageanchor=true}

\section{Introduction}

In a companion paper [2], we introduced a training-free method for dysarthria severity assessment. The present paper provides a large-scale external validation and phenotyping extension of that method, rather than a new representation-learning approach. The method measures the degradation of phonological feature subspaces in frozen HuBERT representations. By ``collapse'' we mean decreased d-prime separation between phonological category pairs (e.g., nasal vs. oral stop) along learned feature directions---not geometric rank collapse or dimensionality reduction. The method computes d-prime (d') scores along phonological contrast directions---nasality, voicing, stridency, sonorance, manner of articulation, and four vowel features---derived exclusively from healthy control speech via Montreal Forced Aligner. Evaluating 890 speakers across 10 corpora and 5 languages, that study demonstrated significant severity correlations for all five consonant d-prime features (pooled Spearman rho = -0.47 to -0.55), with the effect surviving bootstrap confidence intervals, multiple-comparison correction, leave-one-corpus-out sensitivity analysis, and random-effects meta-analysis. The method required no labelled dysarthric training data, generalised across English, Spanish, Dutch, Mandarin, and French despite HuBERT being pre-trained on English only, and produced clinically interpretable 12-dimensional phonological profiles.

However, that initial validation left two clinically important questions unresolved. First, although the method was shown to correlate with overall severity, it was not established whether the pattern of phonological degradation---which features collapse first, and by how much---differs systematically across aetiologies. Clinical dysarthria taxonomy distinguishes hypokinetic (Parkinson's disease), spastic (cerebral palsy, stroke), flaccid (bulbar ALS), and mixed subtypes [1], each affecting different neuromuscular subsystems. Because aetiologies differ in typical severity distributions and recording protocols, observed profile differences could partly reflect severity mix or protocol effects rather than aetiology itself. We address this via severity-matched analyses (Section 4.1), although the strength of this control depends on the validity of the cross-dataset severity harmonisation (see Section 5.1). Two limits of that control are stated at the outset. Within recorded severity categories, feature differences persist among mild speakers, but the consonant features show little separation among moderate speakers pooled across corpora. The apparent Parkinson's profile was sensitive to the healthy-control reference and could not be attributed to aetiology alone, because within recorded severity categories the PD speakers come predominantly from different languages and corpora than the other groups (Section 4.1). If phonological profiles are aetiology-associated, the method could inform differential characterisation: the feature-extraction stage requires no aetiology labels, although any downstream screening would still require cohort-derived aetiological templates. Second, the cross-lingual generality of the method was demonstrated for only 5 languages using a single SSL backbone. It remained unclear whether the findings would replicate across a broader typological sample, including tonal (Mandarin), agglutinative (Swahili, Hungarian, Tamil), and additional Indo-European languages, or whether the results were contingent on the specific representational geometry of HuBERT-base.

A further methodological concern warranted resolution. Our earlier paper identified a token count confound: d-prime values correlate with the number of phone tokens available per speaker, and severely dysarthric speakers tend to produce fewer tokens. While within-corpus analyses and partial correlations suggested the severity signal was genuine, the confound could not be fully eliminated at the scale of 890 speakers distributed across 10 heterogeneous corpora. In the present study (3,374 speakers across 25 datasets), group-level means produce monotonic severity gradients, after regressing out token count the aetiology ordering is preserved except that cerebral palsy and stroke, whose composite means are nearly equal, exchange places, and the severity and aetiology associations persist under fixed token budgets in eligible speakers (Section 4.5). A larger sample does not by itself remove the confound.

The present study addresses these gaps. We scale the phonological subspace analysis pipeline from 890 to 3,374 speakers, from 10 to 25 datasets, and from 5 to 12 languages, adding Slovak, Portuguese, German, Hungarian, Italian, Tamil, and Swahili to the original English, Spanish, Dutch, Mandarin, and French. We extract phonological profiles using 6 SSL backbones spanning three architectural families: the HuBERT family (HuBERT-base, HuBERT-large, WavLM-base), the wav2vec2 family (wav2vec2-base), and multilingual models (XLS-R-300M trained on 128 languages, MMS-300M trained on 1,400+ languages). We evaluate aetiology discrimination across 5 main clinical categories (plus healthy controls) using Kruskal-Wallis tests with epsilon-squared effect sizes and pairwise Cohen's d, cross-lingual profile-shape stability via cosine similarity of aetiology-group profiles, and cross-model robustness via inter-model Spearman correlations and profile cosine similarity.

Our contributions are:

\begin{enumerate}
\def\labelenumi{(\alph{enumi})}
\item
  Aetiology-associated phonological profiles. We show that, at the group level, the pattern of phonological degradation differs across the aetiology groups, with 10 of 13 features yielding large effect sizes (epsilon-squared > 0.14) in distinguishing healthy controls, Parkinson's disease, cerebral palsy, ALS, Down syndrome, and stroke. Parkinson's disease (hypokinetic dysarthria, basal ganglia) is separable from the articulatory execution group (cerebral palsy, Down syndrome, stroke) at a mean pairwise Cohen's d of 1.01 (composite consonant d-prime, speakers with >=3 valid consonant features), while cerebral palsy and stroke are nearly indistinguishable (d = 0.04). A separate moderate-only comparison of PD with the pooled execution group gives d = 0.01 (n = 42 PD, 45 pooled). This is a different statistic from the full-sample mean of three pairwise effects, so the two are not directly comparable, but the moderate-only result indicates that part of the full-sample PD separation may reflect differences in severity mix. Within recorded severity categories, feature differences persist among mild speakers and largely disappear for the consonant features among moderate speakers pooled across corpora, and the apparent Parkinson's profile was sensitive to the healthy-control reference and could not be attributed to aetiology alone (Section 4.1).
\item
  Cross-lingual profile-shape stability. We show that the shape of each aetiology's degradation profile is cross-lingually stable: mean cosine similarity of 5-dimensional consonant d-prime profiles exceeds 0.95 for Parkinson's disease across 6 languages with n$\geq$3 speakers, cerebral palsy across 4 languages, and ALS across 4 languages. For each language pair we bootstrapped the cosine (1,000 resamples of speakers within each of the two languages); averaging the resulting 95\% percentile bounds over the language pairs gives PD 0.978 {[}0.956, 0.984{]}, CP 0.987 {[}0.976, 0.994{]}, ALS 0.985 {[}0.954, 0.989{]} and HC 0.980 {[}0.968, 0.986{]}. These bracketed values summarise the uncertainty of the individual language pairs; they are not a confidence interval for the mean cosine. The lowest PD language-pair cosine is 0.937, and the pair with the lowest bootstrap bound, Dutch--Portuguese PD (cosine 0.968), has a lower bound of 0.893. However, a prespecified direct test of whether this stability is aetiology-specific did not meet its criterion: profiles from the same aetiology in different languages were not detectably more similar than profiles from different aetiologies (block permutation p = 0.055 on the frozen values and p = 0.032 with the documented Tamil direction construction; Delta = +0.0004, 95\% bootstrap CI [-0.003, +0.004]; leave-one-language-out robustness not met in either state; Section 4.2). The result is inconclusive; it does not establish that aetiology-specific cross-lingual structure is absent; it means the observed stability cannot be attributed to aetiology on these data. The constructive cross-lingual finding is at the level of individual contrasts: all five consonant effects replicate in sign across every available language for cerebral palsy (14/15 confidence intervals excluding zero) and Parkinson's disease (all 30 sign-consistent), while ALS is inconclusive (the German cell, n = 5, reverses sign).
\item
  Multi-backbone robustness. We validate that phonological subspace analysis is not contingent on HuBERT-base. All 6 SSL backbones tested produce monotonic severity gradients on point estimates (the mild--moderate margin is not resolved for XLS-R and MMS), inter-model agreement on per-speaker composite consonant d-prime (at least three of five consonant features valid) is rho = 0.83-0.99 across all 15 pairs (n = 3,101 per pair), and the minimum aetiology-profile cosine over the 15 backbone pairs is 0.98 (PD), 0.96 (CP), 0.91 (ALS), 0.96 (DS) and 0.94 (stroke). The HuBERT-base, HuBERT-large and WavLM family clusters at rho > 0.96. Robustness holds at the composite level; the fine-grained ordering of which features collapse most replicates only partially across backbones (Kendall's W = 0.33--0.71).
\end{enumerate}

\section{Related Work}

\subsection{Phonological subspace analysis for severity assessment}\label{phonological-subspace-analysis-for-severity-assessment}

Muller et al. [2] introduced the d-prime phonological profiling method used in this study, demonstrating that frozen HuBERT representations encode phonological contrasts in linearly separable subspaces whose separation degrades monotonically with clinical dysarthria severity. That study validated the approach on 890 speakers across 10 corpora and 5 languages, reporting pooled Spearman correlations of rho = -0.47 to -0.55 between consonant d-prime features and clinical severity, with effects surviving multiple-comparison correction and leave-one-corpus-out analysis. The theoretical foundation draws on Choi et al. [3], who showed that HuBERT encodes phonological features in position-dependent orthogonal subspaces, and Cho et al. [4], who demonstrated strong correlations (r = 0.81) between SSL speech features and electromagnetic articulography measurements. That companion paper is, at the time of writing, a preprint under review (PLOS Digital Health) and has not yet completed peer review; the present study is therefore also a large-scale external check of its method. The present study extends this work from severity correlation to aetiology discrimination, cross-lingual profile-shape stability, and multi-backbone validation.

\subsection{Scalar severity metrics and ASR-confidence baselines}\label{scalar-severity-metrics-and-asr-confidence-baselines}

An alternative approach to training-free severity assessment is to derive a single scalar metric from ASR model behaviour. Halpern et al. [5] proposed PathBench, a benchmark for pathological speech that includes an articulatory precision score (ArtP) computed as the mean frame-level CTC posterior confidence of a wav2vec2-XLSR model. PathBench defines ArtP as a transcript-aligned phonetic scoring metric requiring forced alignment with a reference transcription, and DArtP as the top-performing reference-free variant. Our implementation is neither; it is an approximate ASR-confidence baseline that offers simplicity -- one number per utterance, no transcript-level forced alignment required (unlike PathBench ArtP; note that our main d-prime pipeline still requires MFA phone alignment, Section 3.1) -- but its scalar aggregate (henceforth CTC-Conf, matching Section 4.4) correlates with both prosodic and segmental measures: in our data (Section 4.4) its strongest correlate is speech rate, one of the three prosodic features (rho = +0.501), followed closely by the consonant d-prime features (sonorance +0.487, manner +0.435, stridency +0.418, nasality +0.404, voicing +0.296) and then by the other two prosodic features, vowel duration CV (-0.398) and pause rate (-0.353). In a 10-fold Ridge regression predicting severity from our 13 features, adding CTC-Conf as a 14th feature yields a 6\% RMSE improvement (0.488 to 0.459) with essentially no change in the severity ranking correlation (rho = 0.633 vs 0.632). This suggests that scalar CTC-confidence metrics and multi-dimensional phonological profiles capture complementary but partially overlapping aspects of speech degradation, with the multi-feature approach providing the aetiology-level resolution that a single scalar cannot.

In contrast to the training-free SSL-subspace approach studied here, the dominant clinical-acoustics tradition builds on handcrafted features: openSMILE/eGeMAPS-style functionals, perturbation and voice-quality measures (jitter, shimmer, HNR), formant-based vowel-space metrics, and articulation-rate statistics, either interpreted directly or fed to classifiers (the SALR system of Kadirvelu et al. [6] is benchmarked against such baselines). The clearest bridge between the traditions is Joshy and Rajan [7], who evaluate dysarthria severity classification on UA-Speech and TORGO under a single deep-learning family, comparing DNN, CNN, gated recurrent units, and long short-term memory networks on MFCC and constant-Q cepstral features. They further compare disorder-specific features from prosody, articulation, phonation, and glottal functioning against these baselines, and evaluate low-dimensional i-vector representations classified by a DNN. Their best system -- DNN with MFCC-based i-vectors -- reaches 93.97\% speaker-dependent and 49.22\% speaker-independent accuracy on UA-Speech. Their design compares acoustic representations under a suite of severity-trained classifiers, whereas the subspace method here measures phonological separability directly within a frozen SSL representation, so the two approaches sit on different sides of the representation-versus-classifier boundary rather than being directly comparable on classifier accuracy. Handcrafted features remain attractive for interpretability and low compute, but each targets a single predefined acoustic phenomenon; the two approaches are complementary rather than competing, and a systematic head-to-head on shared corpora remains future work.

\subsection{Cross-lingual dysarthria in SSL embeddings}\label{cross-lingual-dysarthria-in-ssl-embeddings}

Hernandez et al. [8] examined cross-lingual transfer of Parkinson's disease detection using SSL embeddings, training classifiers on one language and evaluating on another. Their work demonstrated that PD-related speech patterns transfer across languages in wav2vec2 and HuBERT representations, but framed the task as binary PD detection rather than multi-aetiology profiling. Rios-Urrego et al. [9] provided clinical evidence that dysarthric speech features in Parkinson's disease show more similarities than differences across languages, supporting the hypothesis tested computationally in the present study. Yeo et al. [10] evaluated cross-lingual severity estimation using three phoneme-level metrics derived from universal phone recognition: phoneme error rate (PER), phonological feature error rate (PFER), and phoneme coverage (PhonCov). Their PhonCov metric -- which measures the loss of phonemic contrasts in ASR output -- is conceptually related to our d-prime measure, but operates on decoder output rather than encoder representations and produces a single severity score rather than a decomposed aetiology-informative profile.

\subsection{SSL representations for pathological speech}\label{ssl-representations-for-pathological-speech}

Self-supervised speech models have become the dominant feature extraction backbone for pathological speech tasks. Baevski et al. [11] introduced wav2vec2, whose representations underpin both the SALR severity classifier ([6]; 70.48\% accuracy on UA-Speech) and the XLS-R multilingual model [12] used in our multi-backbone analysis. Violeta et al. [13] systematically evaluated SSL representations for multiple speech disorder tasks, finding that layer selection and pooling strategy significantly affect downstream performance. Sapkota et al. [14] conducted layer-wise probing of HuBERT-large (24 encoder layers) for severity classification, demonstrating that its later layers (13-15, at roughly two-thirds depth) are most informative, consistent with our use of the final hidden layer. Bae et al. [15] proposed a three-stage framework using pseudo-label augmentation with Whisper-based contrastive learning, achieving SRCC 0.761 across five unseen datasets spanning multiple aetiologies and languages; however, this approach requires labelled training data and model fine-tuning. Javanmardi et al. [16] further demonstrated that pretrained model embeddings outperform conventional acoustic features for both dysarthria detection and severity classification. These studies all employ SSL representations as input features for supervised downstream classifiers. Our approach differs fundamentally: we analyse the representational structure of the SSL model itself, measuring how well it separates phonological categories, rather than training a classifier on top of the representations. This distinction is what enables training-free operation and cross-lingual generalisation without labelled dysarthric data.

\section{Method}

\subsection{Phonological subspace d-prime pipeline}\label{phonological-subspace-d-prime-pipeline}

We apply the phonological subspace analysis pipeline introduced in Muller et al. [2] with the language-specific feature classes described below. Briefly, the method proceeds in four stages. First, speech recordings are force-aligned at the phone level using Montreal Forced Aligner (MFA; McAuliffe et al., 2017 [17]) with language-specific pretrained acoustic models and pronunciation dictionaries (english\_mfa, dutch\_cv, spanish\_mfa, french\_mfa, mandarin\_mfa, italian\_cv, czech\_mfa for Slovak, hungarian\_cv, portuguese\_mfa, german\_mfa, swahili\_mfa, and tamil\_cv; all from MFA 3.x model repository). Second, frame-level embeddings are extracted from frozen HuBERT-base [18] and averaged over each phone interval to produce phone-level mean embeddings. Third, for each phonological binary feature (e.g., {[}+nasal{]} vs {[}-nasal{]}), a feature direction vector is computed as the difference between the mean embeddings of the two phone categories, estimated exclusively from healthy control speakers within the same language. Fourth, each speaker's phone embeddings are projected onto these feature directions, and the separation between the two categories is quantified using d-prime. A minimum of 5 tokens per class is required for a valid d-prime estimate; speakers with fewer tokens for a given contrast receive a missing value for that feature. We quantify category separation with d', the standard signal-detection sensitivity index [19].

The pipeline yields 9 d-prime features per speaker: 5 consonant contrasts (nasality, voicing, sonorance, stridency, manner of articulation) and 4 vowel contrasts (height, lowness, backness, rounding). We supplement these with 3 structural metrics computed from the same embeddings: boundary sharpness (mean cosine distance between adjacent phone embeddings at phone boundaries), cross-position cosine similarity (mean pairwise cosine similarity of embeddings for the same phone across different utterance positions), and vowel triangle area (the area of the triangle formed by mean embeddings of the three corner vowels /a/, /i/, /u/ in HuBERT space). Additionally, 3 prosodic features are computed from MFA TextGrid timing information: speech rate (phones per second of speech), pause rate (proportion of inter-word intervals exceeding 150 ms), and vowel duration coefficient of variation. For the main aetiology discrimination and cross-lingual analyses, we use a 13-dimensional subset excluding boundary sharpness and cross-position cosine similarity, which are content-type dependent (see Section 5.3). The full 15-dimensional profile (9 segmental d-primes + 3 structural + 3 prosodic) characterises each speaker's phonological integrity across articulatory subsystems. Phone-to-feature mappings are defined in language-specific JSON configuration files (one per language, available in the repository) specifying which IPA symbols belong to each binary feature class (e.g., {[}+nasal{]} = \{m, n, ng\} vs {[}-nasal{]} = \{p, b, t, d, k, g\}). Vowel triangle corners default to /a, i, u/ but are overridden for languages where these are not canonical among our 12 (e.g., Slovak, where the corner-vowel inventory adopted in the released configuration differs). Tone and vowel length are not modelled as separate features in the current implementation. For full methodological details, including the mathematical formulation of d-prime in embedding space, feature direction estimation, and alignment quality controls, see Muller et al. [2].

\textbf{Feature direction construction.} Feature directions are estimated per language by pooling all phone tokens from all healthy-control speakers of that language's control corpora, computing the difference of class mean embeddings, and L2-normalising; a direction is estimated only when both classes contain at least five tokens. Tokens are not reweighted by speaker, corpus, or phone identity, so each corpus and phone type influences the direction in proportion to its token count -- in English, LibriSpeech contributes 71.5\% of control tokens (327,580 of 457,850). Section 4.5 reports the sensitivity of the results to equal-speaker, equal-corpus, and equal-phone reweighting, leave-one-control-corpus-out estimation, and cross-corpus direction transfer. In the five non-HuBERT-base backbone result files, 18 SLR65 Tamil control speakers appear twice: a later single-dataset extraction run re-scored them and appended new rows, and its merge step removed earlier rows only for the requested dataset. The two copies differ only in the direction-dependent d-prime columns; the earlier copy used Tamil directions estimated from 21,710 SLR65 control tokens, the later one directions from 164,200 SLR65 and SSNCE control tokens as described above, and the later rows are used; the provenance record (computed from the extraction logs and result files) is in the verification package. In the HuBERT-base master table, two sets of values exist for these 18 SLR65 speakers. The primary analysis uses the earlier set, which was computed before any SSNCE Tamil speaker had HuBERT-base d-prime values; the extraction log of that run was not retained, so its Tamil direction pool is not documented. The later set comes from a HuBERT-base run whose log records Tamil directions from 164,200 SLR65 and SSNCE control tokens, as described above. Section 4.2 reports the effect of using the later set.

\textbf{Feature definitions and coverage.} Each of the five consonant features is a binary contrast between two explicit phone lists, not a full phonological taxonomy. For English: nasal = \{m, n, ng\} vs \{p, b, t, d, k, g\}; voicing = voiced obstruents and sonorants vs voiceless obstruents; stridency = sibilants, labiodentals and affricates vs non-strident stops and sonorants; sonorance = nasals, liquids and glides vs obstruents; and the feature labelled ``manner'' is an occlusion contrast -- stops and nasals (complete oral or nasal closure) vs fricatives -- with affricates, liquids and glides excluded from both classes. Per-language class lists are given in the released configuration files; two languages deviate from the shared template (in Slovak the occlusion contrast is stops and affricates vs fricatives, without nasals; in Hungarian nasals fall in the negative class), which limits the strict comparability of this one dimension for those languages. Place of articulation is not represented in any feature, so the five-dimensional profile measures five specific contrasts rather than a complete articulatory profile.

\textbf{Statistical conventions.} Unless stated otherwise: ``composite consonant d-prime'' is the mean of the five consonant d-prime values over the features available for a speaker, computed for speakers with at least three of the five features valid (n = 3,101 for the cross-backbone Spearman comparison); pairwise Cohen's d uses the pooled standard deviation; the severity correlation is a Spearman correlation over the ordinal scale control = 0, mild = 1, moderate = 2, severe = 3, computed over the six analysis groups including controls; cross-backbone per-speaker agreement is reported on the 5-consonant composite (nasality, voicing, sonorance, stridency, manner); Holm correction is applied over the family of 15 pairwise Mann-Whitney U tests on composite consonant d-prime; and bootstrap intervals are percentile intervals with fixed seeds (1,000--10,000 resamples), over speakers unless an analysis states another resampling unit (the minimum-n and minimum-HC cross-lingual analyses of Section 4.5 resample language pairs, and the cross-lingual summary in contribution (b) averages per-language-pair speaker-bootstrap bounds). The analysis plan designates the block permutation test as primary for the aetiology-specificity comparison (Section 4.2); the bootstrap interval was not part of the analysis plan and is reported alongside the permutation p-value as a secondary summary, not in place of it. The requirement of at least three healthy controls per language for HC-normalised cells, the raw-profile and two-stage-bootstrap sensitivity analyses, and the within-language and raw-profile permutation variants reported in Section 4.2 were defined after the analysis plan was frozen and are post hoc sensitivity analyses rather than prespecified ones.

\textbf{Data corrections.} An audit of the analysis inputs after initial submission found four data errors, which are corrected throughout and disclosed here in full. (i) Twenty-five MDSC control speakers were labelled cerebral palsy in the speaker inventory and in the inputs to all five non-HuBERT-base backbone runs (the HuBERT-base master file carried the correct labels); the error also affected Table 3, the Section 4.1 group counts and Figure 1 as originally submitted; all are reported here from the corrected inputs, and all cross-backbone analyses in Section 4.3 use the corrected labels. (ii) In COPAS, 115 speakers with non-control diagnoses (cleft palate 37, unlabelled 32, laryngectomy 30, mixed 9, voice disorder 7) were flagged as controls, so the submitted Dutch feature directions were estimated from a partially pathological pool; the Dutch directions were rebuilt from the 46 healthy COPAS controls only (the 20 controls carrying mild or moderate labels are healthy by diagnosis, verified against the COPAS manifest; the \texttt{severity\_label} field reflects DIA intelligibility scoring, not clinical diagnosis) and all 254 Dutch speakers with aligned data were re-scored (29 Dutch speakers without recoverable alignments are excluded from the corrected analyses, and 23 of the 254 have too few tokens for any d-prime estimate); Figure 3, which plots Dutch PD profiles, is regenerated from the re-scored values. (iii) The five non-HuBERT-base backbone files split SAP into two dataset labels, and the exact join used for cross-backbone comparison silently dropped 1,045 of 1,233 SAP speakers; the corrected analyses use a canonicalised join. The restored SAP rows (and one Hungarian control) carry no aetiology label in those files, so every analysis takes speaker metadata (aetiology, control status and severity label) from the master table for every backbone. (iv) VOC-ALS appears as an Italian control source in the extraction code's configuration but has no aligned data and no rows in the master table; it contributed nothing to any result and is removed from the dataset accounting. Every number in this paper is reported from the corrected inputs. The direction-sensitivity and degradation/random-direction controls in Section 4.5 re-score cached phone embeddings, which the corrections do not change, with control pools equal to the corrected healthy-control set; their summaries are computed with the corrected labels. The Dutch direction values used by the five non-HuBERT-base backbone runs could not be rebuilt (no cached embeddings for those models); cross-backbone results were rechecked with Dutch speakers excluded: the minimum inter-model Spearman rises from 0.830 to 0.849, and per-aetiology profile-cosine minima change by at most 0.003 (PD 0.980 with and 0.980 without Dutch, HC 0.965 to 0.962; CP, ALS, DS and stroke unchanged), so the unrebuilt Dutch values do not materially affect the reported agreement. This limitation is also stated in Section 4.3.

\subsection{Datasets}\label{datasets}

We compiled a master dataset of 3,374 speakers drawn from 25 datasets spanning 12 languages and 5 main aetiologies. Table 1 summarises the dataset composition. This represents a 3.8-fold increase from the 890 speakers across 10 corpora and 5 languages evaluated in Paper 1 [2].

\textbf{Table 1.} Overview of the 25 datasets spanning 12 languages (26 dataset rows appear because UA-Speech and its healthy-control set are listed separately for readability but count as one dataset). The severity-source column states where each dataset's severity labels come from: a clinical rating (with the instrument where known), intelligibility percentages mapped to classes with the Stipancic thresholds, thresholded listener intelligibility ratings, the corpus documentation, or an assignment by the authors or a collaborator; ``None'' means that the dataset's dysarthric speakers carry no severity label. The evidence for each entry is recorded in the verification package. Language totals count speakers by recording language; some multilingual datasets contribute speakers to multiple languages (e.g., Hungarian Dysarthria includes 8 English-language speakers counted under English), so dataset speaker counts may not sum to the language total.

\begin{longtable}{@{}p{0.21\linewidth}p{0.29\linewidth}p{0.24\linewidth}p{0.22\linewidth}@{}}
\toprule
\textbf{Language (n)}
 & \textbf{Dataset (Speakers)}
 & \textbf{Aetiology}
 & \textbf{Severity source}
 \\
\midrule
English (1,435)
 & SAP [20] (1,233)
 & PD, ALS, CP, DS, Stroke
 & SAP listener intelligibility ratings (1--7), thresholded
 \\
 & LibriSpeech [21] (150)
 & HC
 & N/A
 \\
 & TORGO [22] (15)
 & CP, ALS, HC
 & Corpus documentation
 \\
 & UA-Speech [23] (15)
 & CP
 & Corpus documentation (mild/severe only)
 \\
 & UA-Speech controls [23] (13)
 & HC
 & N/A
 \\
Slovak (602)
 & EWA-DB [24] (602)
 & PD, HC
 & None (PD speakers unlabelled)
 \\
Portuguese (370)
 & AVFAD [25] (370)
 & ALS, PD, HC
 & None (dysarthric speakers unlabelled)
 \\
Dutch (283)
 & COPAS [26] (227)
 & Mixed, cleft palate, laryngectomy, voice disorder, unknown, HC
 & DIA intelligibility \% (Stipancic)
 \\
 & Domotica [26] (43)
 & Mixed
 & None
 \\
 & CHASING [27] (8)
 & PD
 & Listener intelligibility ratings, thresholded
 \\
 & TreasureHunters1 [28] (5)
 & PD
 & Assigned by the authors from published listener results
 \\
Spanish (211)
 & Neurovoz [29] (111)
 & PD, HC
 & Clinical (GRBAS G)
 \\
 & PC-GITA [30] (100)
 & PD, HC
 & Clinical (UPDRS speech item)
 \\
Italian (120)
 & IPVS [31] (65)
 & PD, HC
 & Derived from CPS3 scores
 \\
 & EasyCall [32] (55)
 & Mixed, HC
 & None (dysarthric speakers unlabelled)
 \\
Mandarin (100)
 & MDSC [33] (56)
 & CP, HC
 & Intelligibility \% (Stipancic)
 \\
 & CDSD [34] (44)
 & CP
 & None
 \\
Tamil (80)
 & SLR65 [35] (50)
 & HC
 & N/A
 \\
 & SSNCE (LDC2021S04) (30)
 & CP, HC
 & Corpus documentation
 \\
German (66)
 & SVD [36] (53)
 & HC, unknown, PD, ALS, DS
 & None
 \\
 & YouTube German (collected by authors) (13)
 & ALS, HC
 & Assigned by the authors (not clinical)
 \\
Hungarian (59)
 & Hungarian Dys. [37] (39), CV\_Hungarian (27), Hungarian\_HC (1)
 & PD, Stroke, Mixed, unknown, HC
 & Supplied by a collaborator (15 of 39 speakers)
 \\
French (24)
 & YouTube French (collected by authors) (24)
 & ALS, HC
 & Assigned by the authors (not clinical)
 \\
Swahili (24)
 & CDLI Kenyan [38] (25)
 & CP, PD, MS, HC
 & Corpus documentation
 \\
\bottomrule
\end{longtable}

Of the 3,374 speakers, 2,059 (61.0\%) have severity labels: 1,426 control, 373 mild, 182 moderate, and 78 severe. The remaining 1,315 speakers (39.0\%) have unknown severity, predominantly from the SAP dataset (1,045 speakers without intelligibility ratings) and EWA-DB (78 speakers). Severity labels come from four kinds of source (Table 1 gives the source for each dataset): (i) clinical ratings or severity classes supplied with the dataset (e.g., the UPDRS speech item for PC-GITA, GRBAS G for Neurovoz, the severity classes documented for TORGO and UA-Speech); (ii) Stipancic threshold derivation ([39]; originally developed for ALS but applied here as a pragmatic harmonisation heuristic across aetiologies, on the basis that the thresholds are derived from listener-rated intelligibility percentages, which are aetiology-agnostic) for datasets reporting intelligibility percentages without categorical labels (COPAS, whose percentages come from the Dutch Intelligibility Assessment, and MDSC), using thresholds of >94\% = control, 85-94\% = mild, 70-84\% = moderate, <70\% = severe; (iii) thresholded listener intelligibility ratings (SAP's 1--7 ratings, CHASING's listener ratings) or labels assigned by the authors or supplied by a collaborator (the YouTube recordings, TreasureHunters1 and 15 of the 39 Hungarian Dysarthria speakers); and (iv) healthy control assignment for speakers from normative datasets (LibriSpeech, SLR65 Tamil). A separate cross-check of aetiology labels finds 68 speakers with unknown aetiology and 1 speaker with multiple sclerosis, a known aetiology outside the six analysed here. Not every listed speaker has d-prime features: 186 of the 3,374 speakers have none, including 32 of the 50 SLR65 Tamil control speakers (48 of the 80 Tamil speakers have d-prime features).

MFA alignment was performed using language-specific pretrained acoustic models: english\_mfa (English), spanish\_mfa (Spanish), mandarin\_mfa (Mandarin), french\_mfa (French), dutch\_cv (Dutch), italian\_cv (Italian), czech\_mfa (Slovak), swahili\_mfa (Swahili), and pretrained MFA repository models for Portuguese (portuguese\_mfa), German (german\_mfa) and Hungarian (hungarian\_cv); Tamil uses tamil\_cv from the same repository. Of the 3,374 speakers, 2,961 (87.8\%) yielded a valid nasality estimate; 226 speakers with usable other consonant features but no nasality estimate remain in downstream analyses on the features they support, and 187 speakers have no consonant d-prime (186 have no d-prime at all and 1 has a vowel d-prime only), because they lacked usable MFA alignments or produced too few phone tokens; the 186 come from EasyCall (48), EWA-DB (43), COPAS (32), SLR65 Tamil (32), Domotica (20), Hungarian Dysarthria (8) and three other corpora (1 each).

\subsection{Multi-backbone SSL models}\label{multi-backbone-ssl-models}

To test whether the phonological subspace analysis method is contingent on a specific SSL architecture, we replicated the full pipeline using 6 SSL backbones spanning three architectural families. Table 2 summarises the models.

\textbf{Table 2.} Self-supervised speech models evaluated. All models are used frozen (no fine-tuning). For each backbone, we extract the final hidden layer output. *HuBERT-large-ft was self-supervised pre-trained on 60,000 hours of Libri-Light (hubert-large-ll60k), then fine-tuned on LibriSpeech 960h; we use the fine-tuned checkpoint frozen; WavLM-base [40] was pre-trained on 960 hours of LibriSpeech (WavLM-Large uses 94,000 hours; we use the base variant). The same MFA phone alignments are reused across all backbones.

\begin{longtable}{@{}p{0.16\linewidth}p{0.32\linewidth}p{0.08\linewidth}p{0.20\linewidth}p{0.16\linewidth}@{}}
\toprule
\textbf{Model}
 & \textbf{Identifier}
 & \textbf{Dim}
 & \textbf{Pre-training data}
 & \textbf{Reference}
 \\
\midrule
HuBERT-base
 & facebook/hubert-base-ls960
 & 768
 & English LS 960h
 & Hsu et al., 2021
 \\
HuBERT-large
 & facebook/hubert-large-ls960-ft*
 & 1024
 & English Libri-Light 60Kh (fine-tuned LS 960h)
 & Hsu et al., 2021
 \\
WavLM-base
 & microsoft/wavlm-base
 & 768
 & English LS 960h
 & Chen et al., 2022
 \\
Wav2vec2-base
 & facebook/wav2vec2-base
 & 768
 & English LS 960h
 & Baevski et al., 2020
 \\
XLS-R-300M
 & facebook/wav2vec2-xls-r-300m
 & 1024
 & 128 languages, 436K h
 & Babu et al., 2022
 \\
MMS-300M
 & facebook/mms-300m
 & 1024
 & 1,400+ languages, $\sim$500K h
 & Pratap et al., 2024
 \\
\bottomrule
\end{longtable}

The first four models were pre-trained exclusively on English speech, while XLS-R and MMS were pre-trained on massively multilingual data. All models were used frozen with no fine-tuning. The same MFA phone alignments were reused across all backbones; only the embedding extraction step was repeated. The five non-HuBERT-base backbone result files contain 3,210 rows for 3,192 unique speakers (18 keys are duplicated; Section 3.1). The 182 master speakers absent from them come from EasyCall (47), Hungarian Dysarthria (39), EWA-DB (34), SLR65 Tamil (32), Domotica (20), COPAS (9) and CDLI Kenyan Swahili (1); 151 of these also have no HuBERT-base d-prime, and the other 31 were not processed by the later backbone runs.

\subsection{Token count adjustment}\label{token-count-adjustment}

Muller et al. [2] identified a token count confound in d-prime estimation: speakers who produce more phone tokens tend to yield higher d-prime values, and severely dysarthric speakers tend to produce fewer tokens due to shorter utterances and lower speech rates. At the scale of 890 speakers distributed across 10 heterogeneous corpora, this confound could not be fully eliminated.

A larger sample reduces sampling noise but does not by itself remove a systematic dependence of d-prime on token count. We examined the confound in two ways here and with fixed token budgets in Section 4.5. First, group-level means produce monotonic severity gradients (control > mild > moderate > severe) for all consonant d-prime features. Second, we performed a regression robustness check: we regressed each d-prime feature on n\_phones (linear) and examined the residuals. In the residualised values the severity gradient remains monotonic, and the aetiology ordering is preserved except that cerebral palsy and stroke, whose composite means are nearly equal (d = 0.04), exchange places; the between-group differences in phonological profiles are therefore not explained by a linear dependence on token count. We report raw (unadjusted) d-prime values throughout the main results, with the regression adjustment serving as supplementary validation. This approach is consistent with the recommendation of Westfall and Yarkoni (2016) [41] to report both raw and adjusted effects when the confound is correlated with but not causally responsible for the outcome.

Two further token-count checks are reported in Section 4.5. Token-matched severity comparisons follow the procedure of the submitted analysis, re-run unchanged on the corrected inputs: speakers in the six analysis groups with a recorded severity label (control, mild, moderate or severe) and a composite consonant d-prime are grouped by label; for each pair of adjacent levels, each speaker at the lower level is matched, in table order and without replacement, to the first unmatched speaker at the higher level whose n\_phones ratio lies between 0.8 and 1.25; composite consonant d-prime is compared between the matched groups with a Mann-Whitney U test, and Cohen's d uses the average of the two group variances. Fixed-token estimation re-computes each speaker's d-prime from exactly N tokens per class and scores every subsampled draw twice: along the axis joining that speaker's own two subsampled class means, the estimator of the submitted analysis, and along the healthy-control directions of Section 3.1 with the estimator of the main method. Both scorings use the same draws and, for every language, the phone classes behind the reported per-speaker d-prime values. Its severity correlation uses the recorded severity labels (Section 4.5).

\section{Results}

\subsection{Aetiology discrimination}\label{aetiology-discrimination}

We evaluated whether phonological d-prime profiles differ systematically across aetiologies using Kruskal-Wallis H tests across 6 groups: healthy controls (HC, n = 1,212), Parkinson's disease (PD, n = 616), cerebral palsy (CP, n = 378), amyotrophic lateral sclerosis (ALS, n = 322), Down syndrome (DS, n = 165), and stroke (n = 98). These counts are speakers with a valid nasality d-prime from the HuBERT-base backbone (per-feature N is given in Table 3; the composite-based comparisons below use speakers with at least three valid consonant features, e.g. HC 1,323, PD 638); speakers with missing segmental features or aetiologies outside the main 6 categories (e.g., voice disorder, cleft palate, multiple sclerosis) were excluded from this analysis. Table 3 reports the Kruskal-Wallis H statistic, p-value, and epsilon-squared effect size for each feature. We compute epsilon-squared as (H - k + 1) / (N - k), where H is the Kruskal-Wallis statistic, k the number of groups, and N the total sample size. Effect size thresholds (small < 0.06, medium 0.06-0.14, large > 0.14) follow conventional heuristics adapted from Cohen (1988) [42]. Stratified bootstrap resampling (1,000 iterations, stratified by token-count quartile) yields a composite severity correlation of rho = -0.543 {[}95\% CI: -0.572, -0.512{]} (six analysis groups, n = 1,761), with per-feature CIs ranging from {[}-0.577, -0.520{]} (nasality) to {[}-0.512, -0.443{]} (voicing), none crossing zero. This bootstrap is a re-implementation that codes severity by the Statistical conventions (healthy controls = 0); the original bootstrap script retained with the submission codes each speaker's recorded severity label instead and, on the same inputs, gives rho = -0.542 {[}-0.573, -0.513{]}. We report the re-implementation because it uses the coding of every other severity correlation in the paper; both scripts are in the repository.

\textbf{Table 3. Aetiology discrimination: Kruskal-Wallis H-test across six groups (HC, PD, CP, ALS, DS, Stroke) for all 15 features. The cross-lingual and cross-backbone analyses in Sections 4.2--4.3 use the 13-feature main subset, which excludes boundary sharpness and cross-position cosine similarity (content-type dependent; see Section 5.3). Effect size interpretation: large (epsilon-squared > 0.14), medium (0.06--0.14), small (< 0.06).}

\begin{longtable}{@{}p{0.19\linewidth}p{0.12\linewidth}p{0.19\linewidth}p{0.15\linewidth}p{0.10\linewidth}p{0.17\linewidth}@{}}
\toprule
\textbf{Feature}
 & \textbf{H}
 & \textbf{p}
 & \textbf{epsilon-squared}
 & \textbf{N}
 & \textbf{Effect size}
 \\
\midrule
Nasality d$'$
 & 1,191.4
 & $2.1 \times 10^{-255}$
 & 0.426
 & 2,791
 & Large
 \\
Voicing d$'$
 & 1,253.6
 & $7.4 \times 10^{-269}$
 & 0.424
 & 2,953
 & Large
 \\
Sonorance d$'$
 & 700.3
 & $4.3 \times 10^{-149}$
 & 0.237
 & 2,943
 & Large
 \\
Stridency d$'$
 & 1,239.2
 & $9.4 \times 10^{-266}$
 & 0.431
 & 2,867
 & Large
 \\
Manner d$'$
 & 1,153.7
 & $3.2 \times 10^{-247}$
 & 0.395
 & 2,917
 & Large
 \\
Height d$'$
 & 1,381.6
 & $1.3 \times 10^{-296}$
 & 0.498
 & 2,768
 & Large
 \\
Lowness d$'$
 & 961.7
 & $1.2 \times 10^{-205}$
 & 0.345
 & 2,775
 & Large
 \\
Backness d$'$
 & 1,080.0
 & $2.8 \times 10^{-231}$
 & 0.390
 & 2,760
 & Large
 \\
Rounding d$'$
 & 1,172.2
 & $3.1 \times 10^{-251}$
 & 0.443
 & 2,639
 & Large
 \\
Vowel triangle area
 & 397.4
 & $1.1 \times 10^{-83}$
 & 0.341
 & 1,157
 & Large
 \\
Boundary sharpness
 & 191.7
 & $1.7 \times 10^{-39}$
 & 0.063
 & 2,974
 & Medium
 \\
Cross-position cosine
 & 196.8
 & $1.3 \times 10^{-40}$
 & 0.065
 & 2,974
 & Medium
 \\
Speech rate
 & 155.4
 & $9.2 \times 10^{-32}$
 & 0.055
 & 2,739
 & Small
 \\
Vowel duration CV
 & 278.6
 & $3.9 \times 10^{-58}$
 & 0.105
 & 2,607
 & Medium
 \\
Pause rate
 & 64.6
 & $1.3 \times 10^{-12}$
 & 0.022
 & 2,739
 & Small
 \\
\bottomrule
\end{longtable}

All 15 features discriminate significantly across aetiologies (pause rate is the weakest at p = 1.3 $\times$ 10$^{-12}$; the remaining 14 features are p < 10$^{-30}$). Among the 13 main-analysis features, 10 yield large effect sizes (epsilon-squared > 0.14 by conventional thresholds; Cohen, 1988): the 9 segmental d-prime features and vowel triangle area. The three highest-discriminating segmental features are vowel height d' (epsilon-squared = 0.498), rounding d' (0.443), and stridency d' (0.431). Vowel triangle area (VTA, epsilon-squared = 0.341, Large) is the strongest non-segmental feature and captures global vowel-space geometry in a single scalar, complementary to the individual vowel d-primes; per-aetiology means (HC 31.2, PD 23.7, DS 15.6, ALS 15.5, CP 12.9) show healthy speakers with about twice the triangle area of the ALS and DS groups, 2.4 times that of CP and 1.3 times that of PD and the most severe vowel collapse in cerebral palsy, consistent with strong articulatory involvement in spastic dysarthria. VTA is available for 1,161 speakers (34\%) because it requires at least three tokens each of all three corner vowels /a/, /i/, /u/ per speaker; datasets using controlled word lists, minimal pairs, or sustained phonation often do not meet this requirement. Other non-segmental features show smaller effects: vowel duration CV is medium (epsilon-squared = 0.105), while speech rate and pause rate show small effects (epsilon-squared = 0.022--0.055); boundary sharpness (epsilon-squared = 0.063) and cross-position cosine (0.065) fall just above the medium threshold, indicating that the aetiology signal resides overwhelmingly in the segmental phonological contrasts rather than in prosodic or structural metrics.

Pairwise Cohen's d values on composite consonant d-prime (mean of the 5 consonant features; speakers with at least 3 valid features) reveal a clear clustering structure. Healthy controls are well separated from all dysarthric groups (d = 1.002 vs PD, d = 1.860 vs CP, d = 1.517 vs ALS, d = 1.848 vs DS, d = 1.779 vs Stroke). Critically, PD is separable from all other dysarthric aetiologies: d = 1.014 vs CP, d = 0.617 vs ALS, d = 1.050 vs DS, d = 0.975 vs Stroke. The mean pairwise Cohen's d between PD and the articulatory execution group (CP, DS, Stroke) is 1.013, a large effect. In contrast, CP, DS, and Stroke have very similar composite means: CP vs DS d = 0.012, CP vs Stroke d = 0.036, DS vs Stroke d = 0.061. ALS occupies an intermediate position, closer to the articulatory execution group (d = 0.427 vs CP) than to PD (d = 0.617), consistent with its mixed flaccid-spastic pathophysiology in most patients. Figure 2 shows full-sample group-mean profiles for six aetiology pairs.

This clustering is consistent with known neuroanatomical distinctions, although the matched-severity analyses below indicate that part of it may reflect severity mix and the language and corpus composition of the groups. PD produces hypokinetic dysarthria via basal ganglia dysfunction, characterised by reduced amplitude and velocity of articulatory movements but relatively preserved articulatory targets. CP, DS, and Stroke share severe articulatory impairment affecting similar phonological subsystems, despite differing neurological origins (spastic dysarthria in CP and some strokes, developmental articulatory impairment in DS), all characterised by imprecise consonant production and reduced phonological contrast. The intermediate position of ALS may reflect its dual involvement of both upper and lower motor neurons, with the balance varying across patients and disease stages [1]. The phonological d-prime profiles show group differences consistent with these distinctions without any explicit encoding of aetiology information in the pipeline (Figure 1); they do not by themselves establish a neuroanatomical origin for those differences.

\noindent\makebox[\textwidth][c]{\includegraphics[width=\textwidth,keepaspectratio]{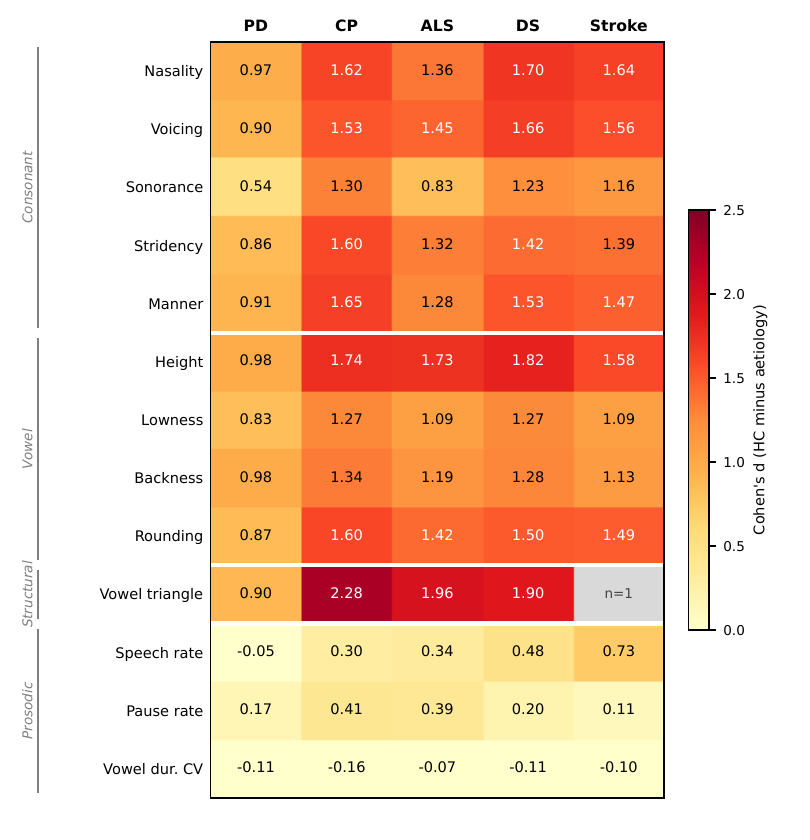}}

\emph{\textbf{Figure 1.} Deviation from healthy controls by aetiology (Cohen's d, healthy controls minus aetiology, pooled standard deviation). Rows show 13 phonological and prosodic features; columns show 5 dysarthric aetiologies. For the d-prime features and vowel triangle area, larger positive values (darker red) indicate a greater reduction relative to healthy controls; for the three prosodic features the value is the healthy-control mean minus the aetiology mean in standardised units. The grey cell (Stroke $\times$ vowel triangle area) is blank because only one stroke speaker has a valid value (a cell needs at least 5).}

To assess how sensitive the aetiology discrimination is to severity distribution differences (e.g., CP having more severe speakers than PD), we ran a moderate-only comparison. Among speakers labelled moderate, the pooled composite consonant d-prime difference between PD (n = 42) and the CP/DS/Stroke articulatory-execution group (n = 45 pooled) is d = 0.01 (Cohen's d = 0.0071). This is a moderate-only restriction, not an equivalence test or full severity matching, and the full-sample d = 1.01 is the mean of three pairwise effects across PD vs. CP, PD vs. DS and PD vs. Stroke rather than a PD-vs-pooled comparison, so the two statistics are not directly comparable. Part of the full-sample PD separation from CP/DS/Stroke may therefore reflect differences in severity mix.

\textbf{Individual features at matched severity.} Because the main claim concerns the pattern of degradation rather than the composite, we repeated the aetiology comparison feature by feature within recorded severity categories. We used the speakers of PD, CP, ALS and DS with a recorded severity label (stroke has only two labelled speakers) and the two strata with enough speakers in more than one group, mild and moderate (severe is almost entirely CP). Within each stratum we ran a Kruskal-Wallis test per feature over the groups with at least 5 valid values, with epsilon-squared as above and Holm correction over the 13 features (Table 4). Among mild speakers (CP 110, PD 98, ALS 68, DS 12), 11 of 13 features differ across aetiologies after correction, six with large effects (height 0.567, rounding 0.539, vowel triangle area 0.540, speech rate 0.225, backness 0.181, voicing 0.172). Among moderate speakers (PD 42, CP 35, ALS 10, DS 9), only height, rounding and speech rate differ, and all five consonant features have epsilon-squared below 0.006 (Holm p = 1), the feature-level counterpart of the moderate-only composite result above.

To ask whether profile shape differs once each speaker's mean feature level is removed, we z-scored each speaker's five consonant d-prime values against all healthy controls and centred them on the speaker's own mean. The statistic was the speaker-weighted between-group sum of squares of the centred group-mean profiles, summed over the two strata, and its null distribution came from 10,000 permutations of aetiology labels within stratum (seed 0). With every speaker referenced to the pooled healthy controls, the centred profiles differ across aetiologies (statistic 51.6 against a null mean of 3.7, p = 0.0001), and the difference is carried by PD: among mild speakers, PD's centred profile has relatively preserved voicing (+0.34) and a relatively weak sonorance contrast (-0.35), whereas CP, ALS and DS show the reverse (voicing -0.15 to -0.28, sonorance +0.22 to +0.37). The pooled controls come mostly from other languages than the PD speakers (1,170 controls, of whom 366 Slovak, 363 Portuguese and 170 English; the matched-severity PD speakers are predominantly Spanish, Italian and Dutch). We therefore repeated the test with each speaker z-scored against the healthy controls of their own language (at least 5 controls; the 12 Swahili speakers, whose language has one control, were dropped). With this reference the PD profile no longer stands out (mild PD centred values within $\pm$0.12 on every contrast), and the centred CP, ALS and DS profiles also change (among mild speakers their sonorance values move from +0.22 to +0.37 to between -0.31 and -0.17). Across aetiologies a shape difference remains (statistic 11.0 against a null mean of 5.6, p = 0.028). It is the change in the centred profiles between the two references, not the omnibus p-value, that shows the profiles depend on the healthy-control reference. The omnibus test shows only that some shape difference across aetiologies remains; it does not show that PD has no distinct profile, and it does not measure the size of the remaining difference. The permutation p-value is (1 + the number of permutations with a statistic at least as large as observed) / 10,001.

This result cannot be attributed to aetiology alone. Within recorded severity categories, the PD speakers come predominantly from the Spanish, Italian and Dutch corpora (97 of 98 mild and 40 of 42 moderate PD speakers). The mild CP, ALS and DS speakers come predominantly from SAP (English); among moderate speakers, all DS and half of the ALS speakers are from SAP, and the CP speakers come from English, Tamil, Swahili and Mandarin corpora. No within-corpus PD comparison meets the minimum of five speakers per group: the only corpora with PD and another aetiology in the same category are SAP (one mild PD speaker) and the Kenyan Swahili corpus (one moderate PD speaker). Severity labels also come from corpus-specific scales (Section 5.1). Restricting the analysis to SAP, where severity, corpus and language are all held fixed but the single mild PD speaker is too few for the comparison, 9 of 13 features differ among mild CP, ALS and DS speakers (82, 64 and 12 speakers; the largest effects are rounding 0.165 and backness 0.152), and the centred profiles still differ (p = 0.0002), but their shapes are close to one another (statistic 2.2 against a null mean of 0.6; with English controls only, 3.6 against 1.8, p = 0.029). The moderate SAP stratum is small (CP 10, DS 9, ALS 5); stridency and manner are the only features significant after correction there. Within recorded severity categories, then, feature-level aetiology differences persist among mild speakers and largely disappear for the consonant features among moderate speakers pooled across corpora. The apparent Parkinson's profile was sensitive to the healthy-control reference and could not be attributed to aetiology alone; because no within-corpus PD comparison meets the minimum of five speakers per group, the language and corpus confound remains. These analyses are post hoc and were not part of the analysis plan.

\textbf{Table 4.} Individual features across aetiologies at matched severity (Kruskal-Wallis epsilon-squared; * Holm-corrected p < 0.05 over the 13 features). All corpora: mild PD 98, CP 110, ALS 68, DS 12 (moderate PD 42, CP 35, ALS 10, DS 9). SAP only: mild CP 82, ALS 64, DS 12. Per-feature n is lower where a feature is missing (vowel triangle area: mild PD 91, CP 38, ALS 49, no DS; moderate PD 35, CP 12, ALS 5, no DS; SAP mild ALS 45, CP 29, no DS).

\begin{longtable}{@{}p{0.30\linewidth}p{0.20\linewidth}p{0.20\linewidth}p{0.20\linewidth}@{}}
\toprule
\textbf{Feature}
 & \textbf{\shortstack{All corpora,\\mild}}
 & \textbf{\shortstack{All corpora,\\moderate}}
 & \textbf{\shortstack{SAP only,\\mild}}
 \\
\midrule
Nasality d' & 0.094* & 0.000 & 0.058* \\
Voicing d' & 0.172* & 0.005 & 0.084* \\
Sonorance d' & 0.022 & 0.003 & 0.092* \\
Stridency d' & 0.112* & 0.000 & 0.088* \\
Manner d' & 0.059* & 0.000 & 0.086* \\
Height d' & 0.567* & 0.183* & 0.053* \\
Lowness d' & 0.112* & 0.000 & 0.082* \\
Backness d' & 0.181* & 0.000 & 0.152* \\
Rounding d' & 0.539* & 0.132* & 0.165* \\
Vowel triangle area & 0.540* & 0.124 & 0.011 \\
Speech rate & 0.225* & 0.234* & 0.038 \\
Pause rate & 0.003 & 0.121 & 0.000 \\
Vowel duration CV & 0.085* & 0.013 & 0.014 \\
\bottomrule
\end{longtable}

\noindent\makebox[\textwidth][c]{\includegraphics[width=\textwidth,keepaspectratio]{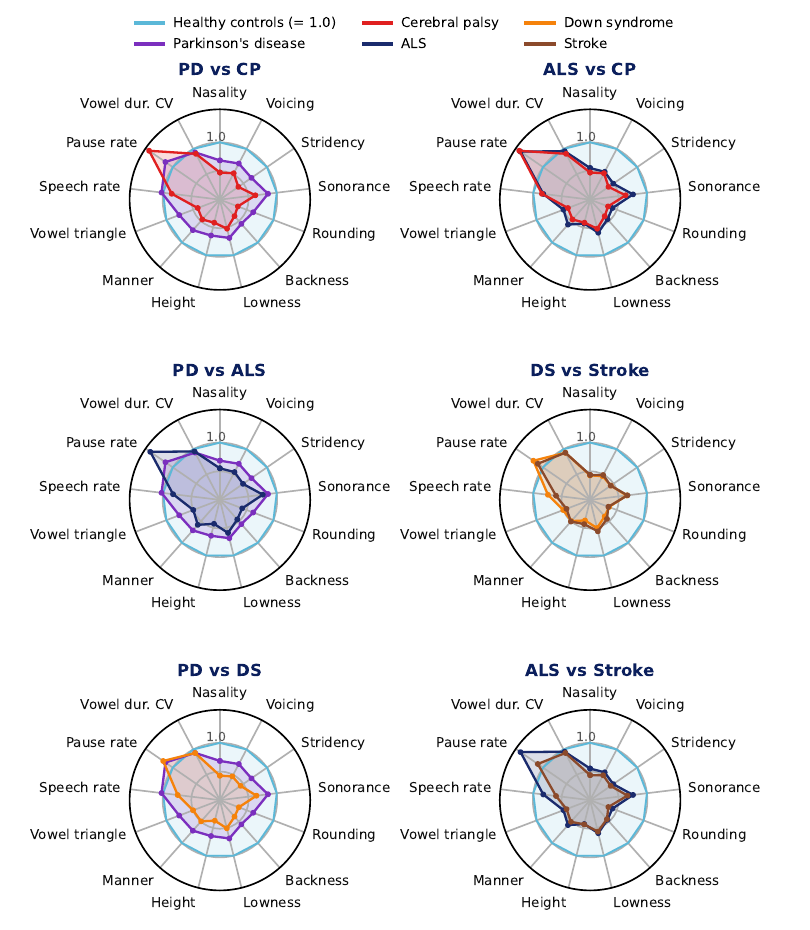}}

\emph{\textbf{Figure 2.} Pairwise aetiology profiles in the full HuBERT-base sample. Panels show group-mean profiles for each pair of aetiologies against the healthy-control reference (light blue, fixed at 1.0). The sample is every speaker in the six analysis groups at all severity levels (HC n = 1,445, PD n = 648, CP n = 379, ALS n = 325, DS n = 166, stroke n = 99), not the moderate-only subset analysed in the preceding paragraph; each feature is the group mean over speakers with a valid value for that feature, divided by the healthy-control mean (inverted for pause rate and vowel-duration CV), so that outward means closer to healthy. The stroke vowel-triangle point rests on the single stroke speaker with a valid vowel triangle area, and the Down syndrome point on 12 speakers.}

\subsection{Cross-lingual profile-shape stability}\label{cross-lingual-profile-shape-stability}

A central question for clinical deployment is whether phonological degradation profiles are language-specific or consistent across languages. If a PD patient in Slovakia shows the same relative pattern of phonological collapse as a PD patient in Brazil, the method can be deployed in new languages without language-specific dysarthric training data, provided healthy-control reference speech and alignment resources are available. We tested this by computing the cosine similarity of mean 5-consonant d-prime profiles (nasality, voicing, sonorance, stridency, manner) for each aetiology across all languages in which that aetiology was represented by at least three speakers with valid features. Table 5 reports the results.

\textbf{Table 5.} Cross-lingual consistency of aetiology-group phonological profiles. Cosine similarity computed on 5-consonant d-prime mean profiles across all languages with at least 3 speakers per cell. The HC row excludes Swahili, whose control set is a single speaker.

\begin{longtable}{@{}p{0.16\linewidth}p{0.24\linewidth}p{0.18\linewidth}p{0.16\linewidth}p{0.16\linewidth}@{}}
\toprule
\textbf{Aetiology}
 & \textbf{Languages}
 & \textbf{Mean cosine}
 & \textbf{Min}
 & \textbf{Max}
 \\
\midrule
CP
 & 4 (en, zh, sw, ta)
 & 0.987
 & 0.974
 & 0.999
 \\
ALS
 & 4 (en, de, fr, pt)
 & 0.985
 & 0.969
 & 0.998
 \\
PD
 & 6 (en, es, it, nl, pt, sk)
 & 0.978
 & 0.937
 & 0.998
 \\
HC
 & 11 (de, en, es, fr, hu, it, nl, pt, sk, ta, zh)
 & 0.980
 & 0.930
 & 0.999
 \\
\bottomrule
\end{longtable}

Mean cosine similarity is 0.978 (PD), 0.987 (CP), 0.985 (ALS), and 0.980 (HC), indicating that the relative pattern of which consonant contrasts are most and least affected is stable across languages. CP shows the highest consistency (0.987 across English, Mandarin, Tamil, and Swahili), followed by ALS (0.985 across English, German, French, and Portuguese) and PD (0.978 across 6 languages with n$\geq$3 speakers). Healthy control profiles also show high cross-lingual consistency (0.980 across 11 languages). Because random directions with no phonological content already give cross-cell cosines near 0.93 (Section 4.5), this high baseline largely reflects the shared geometry of positive-valued d-prime profiles, and it is not evidence that HuBERT encodes phonological contrasts in a language-general manner.

\noindent\makebox[\textwidth][c]{\includegraphics[width=\textwidth,keepaspectratio]{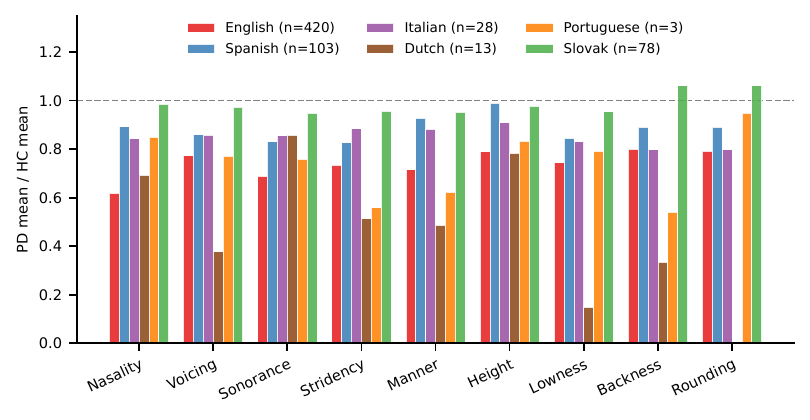}}

\emph{\textbf{Figure 3.} HC-normalised Parkinson's disease profiles in the 6 languages with at least 3 PD speakers. Each bar is the PD mean divided by the same-language healthy-control mean for one of 9 d-prime features (1.0 = healthy). The profiles are broadly similar in English, Spanish, Italian and Slovak; Dutch (n = 13) deviates most, with low voicing and lowness ratios, which may reflect the controlled word lists of the Dutch control material (Section 5.3); the Portuguese bars rest on 3 speakers and Dutch rounding has no valid PD estimate.}

As an anecdotal observation only (n = 1, and the Swahili control baseline is a single speaker), the Swahili PD speaker has a 5-consonant d-prime profile with cosine similarity 0.991 to the multi-language PD mean and 0.988 to the Slovak PD mean, and the 21 Swahili CP speakers show cosine 0.999 with the Mandarin CP mean. Given the result of the aetiology-specificity test below, such similarities should not be read as evidence of aetiology matching in a new language.

\textbf{Is the cross-lingual similarity aetiology-specific?} The within-language permutation analysis in Section 5.1 indicates that the high cross-lingual cosines do not exceed a within-language null. We therefore ran a prespecified direct test of the stronger claim: are profiles from the same aetiology in different languages more similar than profiles from different aetiologies? Over all cross-language cell pairs (cells = language x aetiology with at least 3 speakers, PD/CP/ALS, HC-normalised 5-consonant profiles; HC normalisation additionally requires at least 3 healthy controls in the language, a post hoc rule that excludes the Swahili CP cell because Swahili has a single healthy control), the difference Delta between mean within-aetiology and mean between-aetiology cross-language cosine is Delta = +0.0004 (95\% speaker-bootstrap CI [-0.003, +0.004], 10,000 resamples): mean within-aetiology similarity 0.982 versus between-aetiology 0.982. The bootstrap interval includes zero under a 4-feature profile excluding the occlusion contrast ([-0.005, +0.003]), under the post hoc raw profiles, which include the Swahili CP cell ([-0.004, +0.002]), and under post hoc two-stage dataset-then-speaker resampling ([-0.004, +0.006]), as well as for a 5-speaker cell threshold (Delta = +0.0037, [-0.0005, +0.0106]), and every leave-one-language-out and leave-one-dataset-out deletion. Under centred correlation, which removes the mean profile shape shared by all cells, within-aetiology pairs are in fact slightly less similar than between-aetiology pairs (Delta = -0.066). The analysis plan specified a speaker-level label permutation within language x dataset blocks as the primary inference for this comparison. On the frozen values it gives p = 0.055 (10,000 permutations), so the prespecified criterion for aetiology-specific similarity (Delta > 0 at p < 0.05, with the direction surviving leave-one-language-out and leave-one-dataset-out deletion) is not met; Delta is also negative under raw profiles (-0.0010) and when English (-0.0011), Tamil (-0.0008) or the CDSD corpus (-0.0008) is removed (removing the SSNCE corpus is the same deletion as removing Tamil). The permutation null is thin: of the 19 language x dataset blocks containing PD, CP or ALS speakers, only 5 contain more than one of these aetiologies, and 971 of the 1,010 permutable speakers come from English SAP. Because aetiology is nested within language outside English, no permutation fully preserves the design, and the post hoc permutation variants disagree: permuting within language gives p = 0.016, and the block permutation on raw profiles gives p = 0.001 (raw, within language: p = 0.060). With a 5-speaker cell threshold, the block permutation gives p < 0.001 (2,000 permutations; the leave-one-out deletions were not evaluated at that threshold). We do not read these as evidence of aetiology-specific structure, because they rest on the same few mixed blocks and the bootstrap interval spans zero. The bootstrap interval is a secondary summary; the permutation test is the prespecified primary. The prespecified criterion for aetiology-specific cross-language similarity is therefore not met. This does not establish that aetiology-specific cross-language structure is absent; it means the observed profile-shape stability cannot be attributed to aetiology on these data, and it is consistent with a collapse geometry shared by the aetiologies studied. The primary permutation result is sensitive to how 18 SLR65 Tamil control speakers are scored in the HuBERT-base analysis (Section 3.1). The p = 0.055 above uses the frozen (submission-state) values; re-scoring these speakers with the direction-construction procedure described in Section 3.1, which pools the SLR65 and SSNCE controls, gives Delta = +0.0014 (95\% bootstrap CI [-0.0026, +0.0049]), block permutation p = 0.032 and centred Delta = -0.064; with Portuguese removed, the bootstrap interval then excludes zero ([+0.0002, +0.0072]); and Delta still changes sign when Tamil is removed (-0.0008). Under this re-scoring the primary p-value falls below 0.05, but the bootstrap interval still spans zero and the leave-one-language-out condition of the prespecified criterion is still not satisfied, so the criterion is not met in either state. The result is sensitive and inconclusive, not evidence either for or against an aetiology-specific component.

\textbf{Per-feature cross-lingual replication.} The constructive cross-lingual question is whether individual contrasts are reduced consistently across languages relative to each language's own controls. Using the standardised mean difference of healthy-control minus dysarthric speaker-level d-prime per feature (positive = reduced under dysarthria; bootstrap over speakers, with dataset-then-speaker resampling where a cell or control pool spans multiple corpora): for cerebral palsy, all 5 features are reduced in all 3 languages with usable control baselines (en, ta, zh; 15/15 signs, 14/15 intervals excluding zero; SMD +1.0 to +3.2). For Parkinson's disease, all 5 features are reduced in all 6 languages (30/30 signs), but per-language intervals mostly include zero: Slovak effects are near zero (+0.04 to +0.18) and the Spanish and English intervals are wide under dataset-aware resampling. For ALS the evidence is inconclusive: only Portuguese is individually significant, French is near zero, and the German cell (n = 5: 4 YouTube-sourced speakers and 1 SVD speaker; 4 have a nasality estimate) reverses sign on all 5 features. Cross-lingual replication is therefore strong for CP, directional for PD, and unresolved for ALS.

A critical caveat is that while the profile shape is cross-lingually stable, the absolute d-prime magnitude is not. Kruskal-Wallis tests of raw d-prime across languages give p < 10$^{-9}$ for all nine segmental d-prime features within healthy controls and within PD (across CP and ALS languages, the differences are not significant for every feature), driven by corpus-specific factors including recording conditions, speech task type, and token counts. For example, the COPAS Dutch Intelligibility Assessment uses controlled CVC word lists that inflate d-prime values relative to datasets using spontaneous speech. Additionally, severity labels are not calibrated across datasets: Mandarin severe CP speakers retain only 19-26\% of healthy control d-prime values and English severe CP speakers 30-52\%, while Swahili ``severe'' CP speakers retain 55-74\%, suggesting either different severity assessment scales or unreliable normalisation from a single Swahili control speaker. These findings reinforce our earlier recommendation for Stipancic threshold harmonisation ([39]; originally developed for ALS but applied here as a pragmatic harmonisation heuristic across aetiologies, on the basis that the thresholds are derived from listener-rated intelligibility percentages, which are aetiology-agnostic) and highlight that cross-lingual comparisons should be based on profile shape (cosine similarity) rather than absolute d-prime values.

Data limitations constrain the cross-lingual analysis for two aetiologies. Down syndrome and stroke are represented almost entirely by English SAP speakers (165 of 166 DS and 97 of 99 stroke speakers; the others are 1 German DS speaker in SVD and 2 Hungarian stroke speakers), so no other language has a DS or stroke cell of three or more speakers and neither aetiology can be evaluated cross-lingually. Portuguese and French contribute primarily healthy control speakers (7 and 4 dysarthric speakers respectively), limiting their utility for within-language aetiology comparison. These are data availability constraints, not methodological limitations: the pipeline operates identically for any language with an MFA acoustic model and healthy control reference speakers.

\subsection{Cross-model validation}\label{cross-model-validation}

To test whether phonological subspace analysis is contingent on HuBERT-base, we replicated the full pipeline using 6 SSL backbones and evaluated three aspects of cross-model agreement: severity correlation, inter-model ranking consistency, and aetiology profile similarity.

Severity correlation. All 6 models produce significant negative Spearman correlations between composite consonant d-prime and clinical severity: XLS-R-300M rho = -0.590, HuBERT-base rho = -0.543, MMS-300M rho = -0.541, wav2vec2-base rho = -0.528, WavLM-base rho = -0.527, and HuBERT-large rho = -0.521 (all p < 10$^{-120}$, n = 1,753-1,761; composite over at least three valid consonant features, six analysis groups including controls; speaker metadata from the master table for every backbone). All 6 models produce monotonic severity gradients (control > mild > moderate > severe); the smallest margins are the mild-moderate differences of XLS-R-300M and MMS-300M, whose bootstrap intervals include zero; all other adjacent-severity margins are resolved.

\noindent\makebox[\textwidth][c]{\includegraphics[width=\textwidth,keepaspectratio]{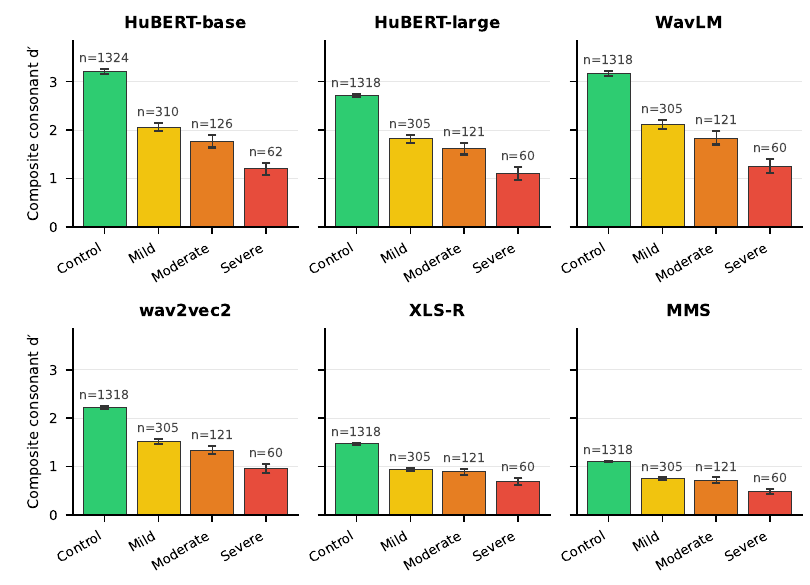}}

\emph{\textbf{Figure 4.} Severity gradient across 6 SSL backbones. Bars show mean composite consonant d-prime (at least three of five consonant features valid) per severity group, counting each speaker once (one row per dataset and speaker identifier). Healthy controls are binned as control regardless of their intelligibility label, and COPAS speakers with non-dysarthric diagnoses (cleft palate, laryngectomy, voice disorder, unlabelled) are excluded from the severity bins; severity labels, control status and aetiology come from the master table for every panel, so all six panels are built the same way. Error bars show 95\% bootstrap confidence intervals of the group mean (1,000 resamples over speakers). The 95\% bootstrap interval of each adjacent-severity difference excludes zero for every step and backbone except the mild--moderate step for XLS-R (difference 0.05, 95\% CI [-0.02, 0.12]) and MMS (0.03 [-0.04, 0.08]). All models show a monotonic decrease from control to severe on point estimates; absolute d-prime magnitudes differ by model architecture.}

\textbf{Table 6.} Inter-model agreement on per-speaker composite consonant d-prime (Spearman rho; at least three of five consonant features valid). All 15 pairs are computed on the same 3,101 speakers. HuBERT-family models cluster at rho > 0.90; multilingual models (XLS-R, MMS) show lower but still strong agreement.

\begin{longtable}{@{}p{0.46\linewidth}p{0.18\linewidth}p{0.18\linewidth}@{}}
\toprule
\textbf{Model pair}
 & \textbf{rho}
 & \textbf{n}
 \\
\midrule
HuBERT-base vs WavLM-base
 & 0.985
 & 3,101
 \\
HuBERT-large vs WavLM-base
 & 0.968
 & 3,101
 \\
HuBERT-base vs HuBERT-large
 & 0.963
 & 3,101
 \\
HuBERT-base vs wav2vec2-base
 & 0.949
 & 3,101
 \\
WavLM-base vs wav2vec2-base
 & 0.948
 & 3,101
 \\
HuBERT-large vs wav2vec2-base
 & 0.934
 & 3,101
 \\
HuBERT-base vs MMS-300M
 & 0.872
 & 3,101
 \\
WavLM-base vs MMS-300M
 & 0.870
 & 3,101
 \\
HuBERT-large vs MMS-300M
 & 0.859
 & 3,101
 \\
HuBERT-base vs XLS-R-300M
 & 0.859
 & 3,101
 \\
WavLM-base vs XLS-R-300M
 & 0.858
 & 3,101
 \\
HuBERT-large vs XLS-R-300M
 & 0.852
 & 3,101
 \\
Wav2vec2-base vs XLS-R-300M
 & 0.839
 & 3,101
 \\
Wav2vec2-base vs MMS-300M
 & 0.834
 & 3,101
 \\
XLS-R-300M vs MMS-300M
 & 0.830
 & 3,101
 \\
\bottomrule
\end{longtable}

Table 6 reveals a clear hierarchical structure. The English-pretrained HuBERT/WavLM/wav2vec2 family clusters at rho = 0.93-0.99, and the multilingual pair (XLS-R vs.\ MMS) falls lowest at rho = 0.83. Across all 15 pairs, computed on the same 3,101 speakers, the range is rho = 0.83 to 0.99. This pattern suggests that phonological subspace structure is a robust property of SSL speech representations in general, not an idiosyncrasy of HuBERT-base.

Aetiology profile consistency. To test whether different backbones produce the same aetiology-specific degradation profiles, we computed cosine similarity of 5-consonant d-prime mean profiles per aetiology for every pair of backbones (15 pairs), with speaker metadata taken from the master table for every backbone (PD 647, CP 378, ALS 325, DS 166 and stroke 97 speakers in the non-HuBERT-base files). With speaker-bootstrap confidence intervals, the per-aetiology minima over the 15 pairs are: PD 0.980 [0.977, 0.984], CP 0.960 [0.954, 0.965], ALS 0.915 [0.908, 0.921], DS 0.955 [0.948, 0.962] and stroke 0.939 [0.925, 0.950] (each for HuBERT-large--MMS), and HC 0.965 [0.962, 0.968] (XLS-R--MMS). While composite-level agreement is high, the fine-grained ordering of which features collapse most agrees only partially across backbones (Kendall's W = 0.44 PD, 0.71 CP, 0.33 ALS, 0.52 DS, 0.48 stroke). Cross-backbone robustness therefore holds at the level of composite severity and group separation, not at the level of exact profile shape, and claims about the exact ranking of features should be read as HuBERT-base-specific.

Leave-one-dataset-out stability. To verify that no single dataset drives the main findings, we repeated the severity correlation and aetiology discrimination analyses after dropping each of the 26 dataset labels in turn (UA-Speech and its control set are dropped separately). Two of these labels (CDSD and Domotica) have no speaker in the severity-labelled six-group sample, so 24 deletions change the sample; all 24 yielded significant severity correlations (rho range: -0.577 to -0.406, mean -0.541; all p < 10$^{-62}$) and large aetiology effect sizes on the composite in that sample (epsilon-squared range: 0.154 to 0.324, mean 0.290; all p < 10$^{-52}$). This analysis is a re-implementation that codes severity by the Statistical conventions (healthy controls = 0); the original leave-one-dataset-out script retained with the submission codes recorded severity labels and gives rho from -0.574 to -0.408 over the same 24 deletions. The most influential dataset is SAP (rho changes from -0.543 to -0.406 when removed), consistent with its dominant sample size; no dataset removal causes a qualitative change in conclusions. Feature importance ranking. Ranking the five consonant contrasts within each backbone by the absolute Spearman correlation of each contrast with severity (the ranking of the submitted analysis, re-run unchanged on the corrected inputs), nasality ranks first in three of six backbones (HuBERT-base, WavLM-base and wav2vec2-base) and is among the top three in four; voicing is among the top three in five, stridency in four, and no contrast is among the top three in all six. The exact feature ordering is therefore backbone-dependent.

\noindent\makebox[\textwidth][c]{\includegraphics[width=\textwidth,keepaspectratio]{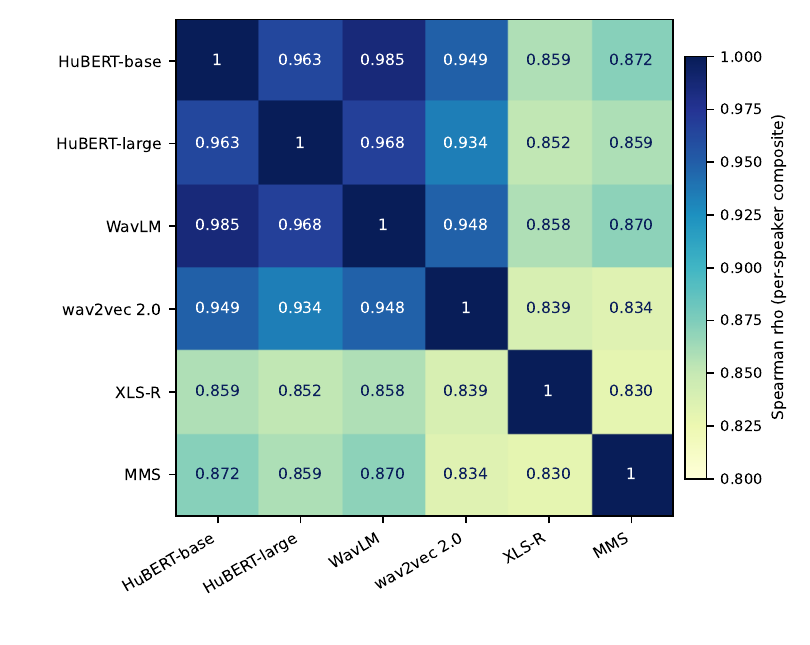}}

\emph{\textbf{Figure 5.} Inter-model agreement on per-speaker composite consonant d-prime (Spearman rho; at least three of five consonant features valid; the same 3,101 speakers for every pair). HuBERT-family models (base, large, WavLM) cluster at rho > 0.90. Multilingual models (XLS-R, MMS) show lower but still strong agreement.}

\subsection{Comparison with ASR-confidence baseline}\label{comparison-with-asr-confidence-baseline}

Halpern et al. [5] proposed PathBench, a benchmark for pathological speech that includes an articulatory precision score (ArtP) computed as the mean frame-level CTC confidence of a language-specific wav2vec2-XLSR model. As an approximate ASR-confidence baseline inspired by but not equivalent to PathBench metrics, we computed a CTC confidence summary (CTC-Conf) for 3,229 speakers across 22 datasets and 12 languages using the language-specific wav2vec2-XLSR-53 CTC models [43] to evaluate whether this simpler metric captures the same information as our multi-dimensional phonological profiles.

CTC-Conf correlates weakly with clinical severity (Spearman rho = -0.190, n = 1,783; six analysis groups, controls = 0, as in the Statistical conventions), substantially weaker than any individual consonant d-prime feature (rho = -0.478 to -0.549). The per-severity mean CTC-Conf values are: control 0.921, mild 0.911, moderate 0.858, severe 0.772. The control-mild separation is small (0.009; Cohen's d = 0.18), limiting CTC-Conf for early-stage severity detection. By contrast, all 5 consonant d-prime features separate control from mild with Cohen's d = 0.97--1.20.

CTC-Conf correlates most strongly with speech rate (rho = +0.501), followed by sonorance d-prime (rho = +0.487), manner (+0.435), stridency (+0.418), nasality (+0.404), vowel duration CV (-0.398), pause rate (-0.353) and voicing d-prime (+0.296). It therefore tracks speech timing and segmental contrast together rather than either alone, consistent with Section 2.2.

In a 10-fold Ridge regression predicting severity (coded as in the Statistical conventions) from our 13 standardised features (n = 806 speakers with all features and CTC-Conf available; ridge penalty 1, seed 42), adding CTC-Conf as a 14th feature yields a modest 6\% RMSE improvement (0.459 vs 0.488) with essentially no change in severity ranking correlation (rho = 0.632 vs 0.633). CTC-Conf alone achieves RMSE = 0.603 and rho = 0.107. These results indicate that CTC-Conf and d-prime profiles capture partially overlapping but largely complementary signals: CTC-Conf is most strongly associated with speech rate but also correlates with the segmental d-prime features, while d-prime profiles measure the separability of specific phonological contrasts. For aetiology-associated characterisation, where the question is not ``how severe'' but ``what pattern of degradation,'' the multi-dimensional d-prime profile provides information that a scalar metric fundamentally cannot.

\subsection{Robustness analyses}\label{robustness-analyses}

We conducted nine robustness analyses to address potential confounds: severity-source ablation, language-minimum-n reruns, Holm-corrected pairwise post hoc, fixed-token d-prime, token-matched severity comparisons, healthy-control baseline sensitivity, SAP-excluded sensitivity, healthy-control direction sensitivity, and generic-degradation / random-direction controls, plus an aetiology-template classifier evaluation.

\textbf{Severity-source ablation.} The master dataset combines clinical severity labels (from published assessments) with healthy-control labels derived from dataset metadata. To test whether results depend on label provenance, we repeated the severity correlation and aetiology discrimination analyses on the clinical-label subset only (n = 705 speakers with clinical severity assessments). The severity correlation remains significant (rho = -0.453, p < $10^{-36}$), and aetiology discrimination yields a large effect (epsilon-squared = 0.188, p < $10^{-27}$). The monotonic severity gradient is preserved: control 2.75, mild 2.09, moderate 1.75, severe 1.15. These values are consistent with the same analysis on all severity-labelled speakers in the six groups (rho = -0.542, epsilon-squared = 0.291; this analysis uses the recorded severity labels, so its rho differs slightly from the -0.543 of Section 4.1, which codes every healthy control as 0), indicating that the findings are not contingent on the inclusion of metadata-derived control labels.

Language minimum-n analysis. To assess whether cross-lingual cosine similarities are inflated by singleton or small-sample language--aetiology cells, we recomputed pairwise cosine similarities with at least 1, 3, 5 and 10 speakers per cell. The intervals in this paragraph are 95\% percentile bootstrap intervals over language pairs (1,000 resamples of the pairwise cosines, seed 42), so they describe the spread of the pairs rather than speaker-level sampling error. For PD, mean cosine similarity is 0.978 {[}0.970, 0.985{]} across 6 languages at n $\geq$ 3 and 0.975 {[}0.965, 0.984{]} across 5 languages at n $\geq$ 10 (5-consonant composite). CP retains cosine 0.987 across the same 4 languages at every threshold ({[}0.980, 0.994{]} at n $\geq$ 3 and n $\geq$ 5, {[}0.980, 0.993{]} at n $\geq$ 10). ALS is represented by only 2 languages (a single language pair) at n $\geq$ 5 and by fewer than 2 at n $\geq$ 10, so the ALS cross-lingual claim should be interpreted cautiously pending larger non-English ALS samples. HC profiles show cosine 0.980 {[}0.975, 0.984{]} across the same 11 languages at n $\geq$ 3, 5 and 10.

\textbf{Multiplicity-corrected post hoc comparisons.} We applied Mann--Whitney U tests with Holm correction across all 15 pairwise aetiology comparisons on composite consonant d-prime. After correction, 12 of 15 pairs remain significant (p < 0.001). The three non-significant pairs are CP vs. DS (p = 1.00), CP vs. Stroke (p = 1.00), and DS vs. Stroke (p = 1.00), consistent with the articulatory execution group interpretation. Signed d values in this section follow the convention that positive d indicates the second-named group has higher composite consonant d-prime (i.e., less collapse); Section 4.1 reports the same magnitudes unsigned. PD is separable from all other groups: PD vs. HC d = +1.00, PD vs. CP d = -1.01, PD vs. ALS d = -0.62. Rank-biserial effect sizes indicate large separation between PD and the articulatory execution group (r = 0.55--0.61) and near-zero separation within it (|r| $\leq$ 0.041).

\textbf{Fixed-token d-prime estimation.} To directly address the token-count confound, we recomputed d-prime from raw phone-level embeddings using a fixed token budget per phonological class. For each speaker and each consonant contrast, we subsampled exactly N tokens per class (discarding speakers with fewer), computed d-prime along the axis joining that speaker's two subsampled class means, repeated 50 times with different random draws, and averaged; a speaker enters a budget with at least three of the five contrasts. Labels come from the corrected inputs. As in the submitted analysis, severity is each speaker's recorded severity label (control, mild, moderate or severe) for speakers in the six analysis groups, rather than the controls-as-0 coding of the Statistical conventions. Table 7 shows that the severity correlation survives at all token budgets from 20 to 200 tokens per class and that the aetiology Kruskal-Wallis test is significant at every budget, but the groups it compares change with the budget. The test uses groups with at least 5 qualifying speakers: stroke never qualifies, so at 20, 50 and 100 tokens it compares HC, PD, CP, ALS and DS, and at 200 tokens PD also drops out (2 qualifying speakers), leaving HC, CP, ALS and DS (N = 465). The PD separation reported in Section 4.1 is therefore not tested at the strictest budget. At the strictest budget (200 tokens), rho = -0.736 (p < $10^{-80}$, n = 467) and epsilon-squared = 0.549. The monotonic severity gradient is preserved at every budget: control > mild > moderate > severe. Within the speakers who meet each budget, the severity association, and the aetiology association among the groups compared, therefore persist under fixed token budgets, although the rising correlation at higher budgets partly reflects a more selective speaker subset (speakers with more tokens tend to have longer recordings) rather than purely improved estimation. To verify this, we repeated the analysis on the 467 speakers who qualify at all token budgets. On this common speaker set, the correlation is stable: rho = -0.750 at 20 tokens, -0.748 at 50, -0.743 at 100, and -0.736 at 200 (all p < $10^{-80}$). On this common set the severity association persists at every fixed token budget, which suggests that the rise at higher budgets mainly reflects which speakers qualify; this comparison does not rule out every effect of the estimator. Because this estimator changes both the token count and the projection direction, we repeated the analysis with the healthy-control directions of the main method (Section 3.1). For each of the same subsampled draws (same speakers, budgets, repeats and random-number sequence), d-prime was computed as in the main analysis: the subsampled tokens were projected onto the language's healthy-control direction for the contrast, and d-prime was the absolute difference of the class means over the root mean of the two class variances. Only the direction and this estimator differ between the two columns of Table 7: the healthy-control column uses the population rather than the sample variance and the absolute rather than the signed difference of the class means. Along the healthy-control directions the severity correlation is rho = -0.589, -0.685, -0.751 and -0.754 at 20, 50, 100 and 200 tokens (all p < $10^{-80}$), the severity gradient is monotonic at every budget (200 tokens: control 3.24, mild 1.86, moderate 1.71, severe 1.20), and the aetiology test is significant at every budget over the same groups (epsilon-squared 0.34--0.59; p < $10^{-55}$). On the 467 speakers common to all budgets, rho is -0.762, -0.761, -0.759 and -0.754. Per speaker, the healthy-control-direction values agree closely with the speaker-own-axis values (Spearman 0.98--0.99) and almost exactly with the healthy-control-direction values computed from all tokens (0.998--1.000), so among the qualifying speakers, with estimates averaged over 50 draws, speaker rankings under the main method's directions are stable across token budgets. The healthy-control directions were rebuilt from the embedding cache and reproduce the reported per-contrast d-prime values to Spearman $\geq$ 0.9999 in all eight languages that qualify. Both columns use the phone classes behind the reported values. An earlier version of the fixed-token script scored Slovak with the German and Swahili with the English phone classes; correcting this changes no conclusion (uncorrected own-axis values: rho = -0.586, -0.672, -0.736 and -0.733), and Table 7 reports the corrected values. The correction also changes which speakers qualify at 20 tokens (854 rather than the 845 of the submitted analysis); within the corrected run, the two columns of Table 7 use identical token draws. Excluding the Slovak and Swahili speakers leaves the result unchanged (rho = -0.566 to -0.763 across budgets and both columns).

\textbf{Table 7.} Fixed-token d-prime: severity correlation and aetiology discrimination at different token budgets per phonological class, with d-prime computed along each speaker's own subsampled class-mean axis (own axis, the estimator of the submitted analysis) and along the healthy-control directions of the main method (HC direction), on the same token draws. The aetiology test compares HC, PD, CP, ALS and DS at 20--100 tokens and HC, CP, ALS and DS at 200 tokens (stroke, and PD at 200 tokens, have fewer than 5 qualifying speakers). Values use the phone classes behind the reported d-prime values for every language (Section 4.5). n is the severity-correlation sample; the 200-token aetiology test uses 465 of these speakers.

\begin{longtable}{@{}llcccc@{}}
\toprule
 & & \multicolumn{2}{c}{\textbf{rho}} & \multicolumn{2}{c}{\textbf{epsilon-squared}} \\
\cmidrule(lr){3-4}\cmidrule(lr){5-6}
\textbf{Tokens/class} & \textbf{n} & \textbf{own axis} & \textbf{HC direction} & \textbf{own axis} & \textbf{HC direction} \\
\midrule
20 & 854 & -0.593 & -0.589 & 0.363 & 0.339 \\
50 & 639 & -0.673 & -0.685 & 0.542 & 0.550 \\
100 & 550 & -0.738 & -0.751 & 0.577 & 0.587 \\
200 & 467 & -0.736 & -0.754 & 0.549 & 0.564 \\
\bottomrule
\end{longtable}

\textbf{Token-matched severity comparisons}. As a complementary check (procedure in Section 3.4), speakers at adjacent severity levels were matched on n\_phones. All three differences are in the expected direction and significant: control vs. mild (n = 185 matched pairs, Cohen's d = 0.60, p < $10^{-6}$), mild vs. moderate (n = 99, d = 0.54, p < $10^{-3}$), and moderate vs. severe (n = 37, d = 0.94, p < $10^{-3}$).

\textbf{Healthy-control baseline sensitivity.} Because feature directions are estimated from healthy controls within each language, the reliability of those estimates depends on HC sample size. We tested whether the cross-lingual cosine results are sensitive to HC adequacy by restricting the analysis to languages with at least 1, 5, 10 or 20 healthy-control speakers (aetiology cells with at least 3 speakers; the intervals are 95\% percentile bootstrap intervals over language pairs, 1,000 resamples of the pairwise cosines). Excluding Swahili (1 HC speaker) at the HC $\geq$ 5 threshold has negligible effect: PD cosine remains 0.978 {[}0.970, 0.985{]} across 6 languages, CP is 0.984 {[}0.974, 0.994{]} across 3 languages (against 0.987 across 4 languages with Swahili), and ALS remains 0.985 {[}0.977, 0.992{]}. The cosines are identical at HC $\geq$ 10 and $\geq$ 20 (only the lower PD bound moves, to 0.969, at HC $\geq$ 20), indicating that the cross-lingual stability finding is not driven by fragile HC baselines.

\textbf{SAP-excluded sensitivity}. Because SAP contributes 1,233 of 3,374 speakers (36.5\%) and nearly all Down syndrome and stroke speakers, we repeated the aetiology and severity analyses with SAP excluded (n = 2,141). The severity correlation holds but is weaker: rho = -0.408 (n = 1,573, p < $10^{-63}$) versus -0.542 with SAP (n = 1,761, p < $10^{-134}$; recorded severity labels, as in the severity-source ablation). Aetiology discrimination on composite consonant d-prime across the 4 groups represented outside SAP (HC/PD/CP/ALS) remains significant but smaller: epsilon-squared = 0.162 (H = 276, p < $10^{-58}$) versus 0.435 for the six groups with SAP. Among the dysarthric groups (five with SAP, PD/CP/ALS without), the consonant effects weaken (e.g., nasality 0.180 $\to$ 0.066), while height and rounding effects are slightly larger without SAP (height 0.190 $\to$ 0.205, rounding 0.165 $\to$ 0.221); because the group sets differ, these per-feature comparisons are descriptive. ALS is thinned to n = 14 outside SAP. These values are a re-analysis on the corrected inputs with the SAP-excluded script in the public repository; on the submitted inputs that script does not reproduce all of the SAP-excluded values reported at submission, so we do not treat it as the script that produced them. SAP contributes 420 PD, 311 ALS, 240 CP, 165 DS and 97 stroke speakers to this analysis. Down syndrome (n = 1) and stroke (n = 2) effectively disappear outside SAP; these aetiology claims are therefore SAP-specific and English-monolingual. The four-group discrimination persists without SAP at a smaller effect size; no cross-corpus robustness claim is made for ALS from this test, because ALS is thinned to n = 14 outside SAP.

\textbf{Healthy-control direction sensitivity.} Because feature directions pool control tokens without reweighting (Section 3.1), we rebuilt all directions under equal-speaker, equal-corpus, and equal-phone-class weighting, under leave-one-control-corpus-out estimation for English, and under cross-corpus transfer (directions from LibriSpeech controls only, and from the clinical corpora only). All variants leave per-speaker composite rankings essentially unchanged: for English, directions estimated without LibriSpeech entirely -- removing 71.5\% of control tokens -- rank speakers at Spearman rho = 0.997 against the published construction, all variants reach at least rho = 0.997 in English and rho = 0.967 in every language tested (cached embeddings re-scored; summaries computed with the corrected labels), and the English group-level effect size varies only in the third decimal (epsilon-squared 0.628-0.632 across ten variants). The construction choice documented in Section 3.1 therefore does not drive the results.

\textbf{Generic-degradation and random-direction controls.} Two controls probe how much of the profile reflects dysarthric phonology as opposed to generic signal degradation or representation geometry. First, applying synthetic degradations to held-out control speech (additive noise at 10 and 0 dB SNR, 3 kHz low-pass, resample-based speed change of +/-15\%) and re-running the identical pipeline reduces d-prime substantially (to 0.45-0.76 of clean values at 0 dB) and produces profile shapes whose similarity to the aetiology profiles is high on the same cosine metrics -- so cross-corpus shape similarity per se cannot be read as evidence of shared phonology. The collapse ordering differs, however: under dysarthria sonorance is the least reduced of the five contrasts in every aetiology group, and nasality is the most reduced by effect size in PD, DS, stroke and the pooled dysarthric group (manner in CP, voicing in ALS), whereas additive noise reduces occlusion (manner) and stridency most. Sonorance is the least-reduced contrast under both dysarthria and noise, so the orderings differ only at the most-reduced end (nasality in most dysarthric groups versus occlusion and stridency under noise); we note this as a possible specificity marker for future work, not as a finding. Second, replacing the phonological directions with random unit directions (matched phone classes and token counts, 50 seeded draws) yields cross-cell profile cosines averaging 0.93, showing that most of the absolute cosine magnitude arises from the geometry of positive-valued d-prime profiles; and random directions already separate controls from dysarthric speakers at roughly half the real effect (English: real SMD 2.72 vs random 95th percentile 1.74). The phonological directions add separation beyond this generic floor where statistical power exists (English clearly; Mandarin marginally), while in the smaller, milder non-English corpora the observed separation does not exceed the random-direction floor.

\textbf{Aetiology-template classification.} To test whether the observed profile differences support individual-level aetiology identification, we evaluated a leave-one-dataset-out nearest-centroid classifier (cosine distance to class centroids of raw 5-dimensional consonant d-prime profiles; speakers with at least 3 of 5 features valid, a missing feature filled with the training-fold feature mean) across 6 groups (2,928 speakers). The classifier achieves 39.8\% overall accuracy, 27.5\% balanced accuracy, and 21.9\% macro F1 (20.5\% when each speaker is compared with the centroids on its available features only). The script that produced the originally submitted value was not retained and could not be recovered; the protocol above was re-implemented and applied to the corrected inputs. HC speakers are best identified (F1 = 0.627), followed by CP (0.388) and ALS (0.231); PD, DS, and Stroke are poorly classified (F1 = 0.000--0.058). The conclusion is insensitive to the protocol: across scaling (raw, z-scored), distance (cosine, Euclidean) and missing-value handling (training-mean fill, comparison on available features), macro F1 ranges from 0.205 to 0.305. While the profiles exhibit statistically robust group-level differences, the within-group variability is too large for template matching to support individual-level aetiology screening; the profiles support population-level phenotyping only.

\section{Discussion}

The results presented above support phonological subspace analysis as a robust framework for aetiology-aware dysarthria characterisation across languages and SSL architectures. In this section, we discuss the interpretation and limitations of these findings.

\subsection{Shape consistency versus magnitude variability}\label{shape-consistency-versus-magnitude-variability}

The cross-lingual analysis reveals a dissociation between two aspects of phonological profiles. The relative pattern of degradation---which features collapse most, and in what order---is consistent across languages. Mean cosine similarity exceeds 0.97 for all aetiology groups tested, and even the single Swahili PD patient aligns with the multi-language PD mean at cosine 0.991, although this observation is anecdotal given n = 1. A permutation analysis qualifies this interpretation: shuffling aetiology labels within languages to create random speaker subsets of the same size yields null cosine distributions comparable to the observed values. For PD, the submitted manuscript reported observed 0.979, null mean 0.982, p = 0.84; the permutation draws behind these values were not retained, and the permutation code retained with the submission does not reproduce them (on the submitted inputs it gives observed 0.979, null mean 0.998, p = 1.00). A seeded reconstruction on the corrected inputs gives observed 0.978, null mean 0.975, p = 0.21. In each case the observed cosine does not exceed the within-language null. This indicates that the high cross-lingual cosine reflects strong within-aetiology profile homogeneity rather than a special cross-lingual signal. The cross-lingual result is therefore best interpreted as profile-shape stability rather than as a demonstration of language independence per se. The prespecified direct test in Section 4.2 is consistent with this reading: its criterion for aetiology-specific cross-language similarity was not met, which does not establish that such structure is absent. The planned block permutation gives p = 0.055 on the frozen values and p = 0.032 when 18 Tamil control speakers are re-scored with the documented direction-construction procedure, while the bootstrap interval spans zero and the leave-one-language-out requirement fails in both states (Section 4.2); the test should therefore be read as sensitive and inconclusive rather than as evidence for or against aetiology-specific structure. The generic-degradation and random-direction controls in Section 4.5 show that high profile-cosine similarity can arise without phonological or pathological signal; this is consistent with a shared profile geometry but does not show that no aetiology-specific component exists. The cross-lingual claims of this paper are accordingly claims about profile-shape stability and per-feature sign replication, not about aetiology-specific cross-lingual structure.

However, the absolute d-prime magnitudes differ substantially across languages and corpora. Kruskal-Wallis tests of raw d-prime across languages give p < 10$^{-9}$ for every segmental feature within healthy controls and within PD. Several factors drive this variability. Recording conditions differ across datasets: clinical recordings in sound-treated rooms (e.g., EWA-DB) yield cleaner signals than home recordings (e.g., YouTube corpora). Token counts vary with speech task length and speaker capacity; as demonstrated in our previous work [2], d-prime values correlate with the number of available phone tokens per speaker. Speech task content also matters: datasets using controlled reading passages or word lists yield different d-prime distributions than those using spontaneous speech.

Most critically, severity labels are not calibrated across datasets. English and Mandarin speakers with severe CP retain 19--26\% (Mandarin) and 30--52\% (English) of healthy-control d-prime values, indicating deeply collapsed phonological contrasts. In contrast, Swahili speakers labelled ``severe'' retain 55--74\% of healthy-control values, more than English CP speakers labelled mild (44--62\%) or moderate (31--48\%). Two explanations are plausible: the CDLI Kenyan Swahili corpus may use a different severity assessment scale from SAP or MDSC. Additionally, the Swahili healthy-control baseline comprises only a single speaker, making HC-normalised ratios for this language unreliable; cross-lingual cosine-similarity results for Swahili should therefore be interpreted with caution. This finding reinforces our recommendation for severity-scale harmonisation using Stipancic thresholds ([39], originally developed for ALS but applied here as a pragmatic harmonisation heuristic across aetiologies because the thresholds are derived from listener-rated intelligibility percentages, which are aetiology-agnostic) and highlights that cross-lingual comparisons must be based on profile shape (cosine similarity or HC-normalised ratios) rather than raw d-prime magnitude.

\subsection{Token count confound: resolution and residual concerns}\label{token-count-confound-resolution-and-residual-concerns}

Our initial work identified the token-count confound as a methodological concern: d-prime values correlate with n\_phones, and severely dysarthric speakers produce fewer tokens. In the present sample, group-level severity means produce monotonic gradients for all consonant d-prime features, and after regressing out n\_phones (linear adjustment) the severity gradient stays monotonic and the aetiology ordering is preserved except that cerebral palsy and stroke, whose composite means are nearly equal, exchange places. The range of group means compresses slightly after adjustment (e.g., the HC--DS composite gap narrows from 1.63 to 1.59).

This apparent attenuation should be qualified. Token count is partly a function of severity itself: severely dysarthric speakers produce fewer phones per utterance because their speech is slower, more effortful, and more frequently interrupted by pauses. Token count is therefore simultaneously a confound (inflating d-prime in speakers with more data) and a signal (reflecting the reduced productive capacity that is itself a symptom of severity). By token-count quartile (composite over at least three valid consonant features, n = 1,761; boundaries computed over the six analysis groups including controls; ordinal severity scale as in Section 3), the severity correlation is Q1 (up to 127 tokens) rho = -0.06, p = 0.105, n = 756, of whom 1 is dysarthric; Q2 (128--183) rho = -0.22, p = 0.016, n = 125, 2 dysarthric; Q3 (185--2,222) rho = -0.39, p < 10$^{-16}$, n = 440, 161 dysarthric; and Q4 (above 2,222) rho = -0.70, p < 10$^{-66}$, n = 440, 274 dysarthric. The two lower quartiles consist almost entirely of healthy controls, so they do not test the severity association at low token counts; the data therefore neither support nor rule out a severity signal in speakers with few tokens. Complete decorrelation is neither achievable nor desirable, as removing all severity-correlated variance would attenuate the very effect of interest. Our approach---reporting raw values as the primary analysis and the regression adjustment as a robustness check---is consistent with the recommendation of Westfall and Yarkoni (2016) [41] for confounds that are partially collinear with the variable of interest.

For individual-level clinical assessment, the token count sensitivity implies that within-corpus calibration remains necessary. A speaker who produces 50 phone tokens in a short recording will have noisier d-prime estimates than one who produces 500 tokens in an extended session, regardless of their actual severity. The fixed-token d-prime experiment (Section 4.5) provides the most direct check: when d-prime is recomputed from exactly N tokens per class using raw phone-level embeddings, the severity correlation persists across token budgets in eligible speakers (rho = -0.750 to -0.736 on a common speaker set of 467 speakers qualifying at all budgets along each speaker's own class-mean axis, and -0.762 to -0.754 along the healthy-control directions of the main analysis), and the aetiology discrimination remains significant at all budgets among the groups with at least 5 qualifying speakers (epsilon-squared 0.36--0.58 along the own axis and 0.34--0.59 along the healthy-control directions; stroke is absent at every budget and PD at 200 tokens). Token-matched severity comparisons further show that all adjacent severity-level differences remain significant when speakers are matched on token count (d = 0.54--0.94). Among the eligible and token-matched speakers, the severity association therefore persists under fixed token budgets, under either projection direction; confidence-weighted d-prime remains a planned extension for deployment scenarios.

\subsection{Content-type effects}\label{content-type-effects}

The COPAS Dutch healthy-control speakers show anomalously high vowel-lowness d-prime (8.54, n = 7 speakers with a lowness estimate, versus a global healthy-control mean of 3.03). Investigation of the COPAS manual (Van Nuffelen et al. [44]) reveals the root cause: the Dutch Intelligibility Assessment (DIA) uses CVC word lists specifically designed to maximally separate phoneme categories, including a subtest targeting all medial vowels and diphthongs of Dutch with 16 controlled items. This concentrated, controlled vowel content produces artificially high d-prime values---the test material is, by design, optimally discriminable.

This content-type confound is not unique to COPAS. Datasets that use controlled reading materials or minimal-pair word lists may inflate d-prime values relative to datasets based on spontaneous speech, because phonological contrasts are more consistently realised in carefully articulated read speech than in casual connected speech. The effect is most pronounced for vowel features, where the difference between controlled and spontaneous contexts is largest. COPAS consonant d-prime values fall within the normal cross-lingual range: for nasality, the COPAS control mean is 3.55 (n = 37 controls with a nasality estimate), the all-speaker mean is 3.03, and the global healthy-control mean is 4.07 -- comparing controls with controls, COPAS runs somewhat lower than the pooled control baseline, so the inflation is a lowness-specific and vowel-specific phenomenon driven by the DIA test design rather than a consonant-side artefact.

For cross-lingual comparisons, the content-type confound means that healthy control baselines are not directly comparable across datasets that use different speech elicitation tasks. HC-normalised ratios partially account for this (because both numerator and denominator are affected), but the degree of inflation may differ between healthy and dysarthric speakers if dysarthric speakers simplify their productions regardless of task demands. We report this as a limitation and note that the cosine similarity analysis, which captures relative profile shape rather than absolute magnitude, is robust to uniform scaling differences across corpora.

\subsection{Data limitations}\label{data-limitations}

Several data gaps constrain the generality of our findings. Down syndrome and stroke are represented almost exclusively by English speakers from the SAP dataset (apart from 1 German DS speaker and 2 Hungarian stroke speakers), precluding cross-lingual validation for these aetiologies. Whether their profiles replicate across languages cannot be assessed until non-English corpora for these aetiologies are available. The Swahili healthy control baseline comprises a single speaker, rendering HC-normalised severity ratios for this language unreliable; Swahili cross-lingual results should be interpreted as preliminary.

The SVD German corpus contributes 53 speakers: 30 healthy controls, 1 PD, 1 ALS, 1 DS and 20 with unknown aetiology. All 53 have voicing, sonorance, stridency, manner, height, backness and rounding d-prime values; nasality d-prime is unavailable for all of them. All 53 meet the at-least-three-feature rule for the composite; the 30 controls and 3 dysarthric speakers enter the six-group analyses, and the 20 speakers with unknown aetiology do not.

Finally, Portuguese and French contribute primarily healthy control speakers (7 and 4 dysarthric speakers respectively), while Tamil and Hungarian contribute limited dysarthric data for within-language aetiology comparison. This limits the cross-lingual aetiology analysis for these languages but does not undermine the positive findings: the languages that do contribute dysarthric speakers across multiple aetiologies support the per-aetiology sign-replication finding heterogeneously: cerebral palsy replicates across English, Mandarin and Tamil (all five contrasts, 14/15 CIs excluding zero); Parkinson's disease is sign-consistent across English, Spanish, Slovak, Italian, Dutch and Portuguese (30/30) with wide per-language intervals in Slovak, Spanish and English; ALS is inconclusive, and the German ALS cell (n = 5) reverses sign on all five features. We therefore make no claim that DS or stroke would replicate cross-lingually pending data.

\subsection{Clinical implications}\label{clinical-implications}

Several properties of phonological subspace analysis make it attractive for clinical translation. First, the method is training-free with respect to pathological speech data: it requires only healthy control recordings in the target language to estimate feature direction vectors, with no labelled dysarthric training data. This dramatically lowers the barrier for deployment in new languages, where dysarthric speech corpora are scarce or nonexistent. A clinician in a language not represented in existing dysarthric speech databases could deploy the system using only a modest healthy control reference set and an MFA pronunciation dictionary.

Second, the method produces interpretable, feature-level output. Rather than a single severity score, the 13-dimensional profile indicates which phonological subsystems are degrading and by how much. This aligns with clinical practice in speech-language pathology, where assessment instruments decompose dysarthria into articulatory, phonatory, resonatory, and prosodic subsystems [1]. At the group level, a clinician-facing summary might report, for example, that a cohort's nasality and occlusion contrasts have degraded more than its sonorance contrast, the pattern observed at the group level for most aetiologies here (Figure 1); individual-level readouts require calibration beyond the current pipeline. Vowel triangle area adds a complementary single-scalar summary of vowel-space geometry; because it shrinks monotonically with dysarthria severity (rho = -0.519) and discriminates aetiologies strongly (epsilon-squared = 0.341), it is a natural candidate for longitudinal monitoring in contexts where within-speaker change over time is clinically informative, provided the recording protocol reliably elicits all three corner vowels.

Third, the group-level aetiology phenotyping demonstrated in Section 4.1 documents population-scale phonological patterns that differ across the aetiology groups studied, within the matched-severity and control-reference limits of that section. This is cohort characterisation, not individual diagnosis. However, a leave-one-dataset-out nearest-centroid classification experiment yields only 21.9\% macro F1 and 27.5\% balanced accuracy across 6 groups, indicating that the observed profile differences, while statistically robust, do not yet support individual-level aetiological screening. The profiles are better suited to population-level phenotyping than to clinical triage at this stage, particularly in settings where specialist neurological assessment is not readily available.

Fourth, the multi-backbone robustness demonstrated in Section 4.3 means that the method is not tied to any specific model release. As SSL speech models evolve, the six-backbone comparison here supports evaluating this framework with other frozen SSL speech models; backbone-specific calibration of absolute d-prime magnitude and clinical validation on the target aetiology and language remain necessary before any new backbone is used clinically.

Finally, profile-shape stability at the group level -- the aspect of cross-language similarity that Section 4.2 supports -- has practical implications for longitudinal monitoring. If the shape of phonological collapse is stable within a language and within a speaker over time, within-speaker changes can be interpreted against that speaker's own baseline rather than against a cross-population template; this is a within-speaker change-detection framing, not an aetiology-template framing (the prespecified aetiology-specific similarity test in Section 4.2 was not met). This is particularly relevant for progressive conditions such as ALS, where tracking the rate and pattern of phonological collapse within a patient could inform clinical decisions about communication augmentation, subject to within-corpus calibration for absolute d-prime magnitude. Longitudinal validation using within-speaker repeated measures is planned future work, pending the availability of ALS longitudinal speech corpora currently being collected.

\section{Conclusion}

This study supports three principal findings. First, phonological subspace analysis in frozen SSL speech representations produces group-level aetiology-associated degradation profiles that discriminate among Parkinson's disease, cerebral palsy, ALS, Down syndrome, stroke, and healthy controls, with 10 of 13 features yielding large Kruskal-Wallis effect sizes (epsilon-squared > 0.14). Holm-corrected pairwise comparisons support that Parkinson's disease is separable from all other groups in the full sample (Cohen's d = 0.62--1.05 in the full sample; a separate moderate-only comparison of PD with the pooled CP/DS/stroke group gives d = 0.01, a different statistic that is not directly comparable), while CP, DS, and Stroke have similar composite means (pairwise d = 0.012--0.061; all pairwise p > 0.5 after correction), which does not by itself show that the three groups are alike. These are population-level distinctions: leave-one-dataset-out template classification reaches only 21.9\% macro F1, so the profiles support population phenotyping, not individual-level diagnosis or screening. Within recorded severity categories, feature differences persist among mild speakers but the consonant features show little separation among moderate speakers pooled across corpora, and the apparent Parkinson's profile was sensitive to the healthy-control reference and could not be attributed to aetiology alone, because aetiology is confounded with language and corpus within recorded severity categories. Second, for the aetiologies with multilingual data (PD, CP, ALS), the relative shape of phonological collapse is stable across languages (mean 5-consonant profile cosine 0.978--0.987), and individual contrasts replicate in sign across languages for CP and PD; the prespecified criterion for aetiology-specific cross-language similarity was not met (block permutation p = 0.055 on the frozen values and p = 0.032 with the documented Tamil direction construction, with the bootstrap interval spanning zero and leave-one-language-out robustness failing in both states), so the result is inconclusive and the stability cannot be attributed to aetiology on these data; this does not establish that aetiology-specific cross-language structure is absent. Third, the method is robust at the composite level across all 6 SSL backbones tested: severity gradients are monotonic on point estimates, inter-model agreement on the 5-consonant composite (at least three features valid) is rho = 0.83-0.99 across all 15 pairs (n = 3,101 per pair; minimum pair XLS-R vs.\ MMS at 0.83), and per-aetiology profile similarity across backbone pairs is at least 0.91; fine-grained feature orderings replicate only partially across backbones.

These findings survive multiple robustness checks: fixed-token d-prime estimation from raw embeddings (rho = -0.762 to -0.754 on a common speaker set across all token budgets, along the healthy-control directions of the main method), severity-source ablation on clinical-only labels (rho = -0.453, epsilon-squared = 0.188), cross-lingual cosine analysis with minimum n $\geq$ 10 per cell (PD cosine 0.975, 5-consonant composite; 5 languages qualifying at n$\geq$10), and token-matched severity comparisons (all p < 0.001). Validated across 3,374 speakers drawn from 25 datasets spanning 12 languages and 5 aetiologies, the results support phonological subspace analysis as a robust, training-free framework for aetiology-aware dysarthria characterisation. The method requires no labelled dysarthric training data, was applied unchanged across 12 languages and six SSL architectures within the limits stated above, and produces clinically interpretable output at the level of individual phonological subsystems. The framework appears sufficiently mature for targeted clinical validation studies, subject to stronger harmonisation and uncertainty analyses to evaluate its utility for severity monitoring, aetiology characterisation, and treatment-outcome measurement in real-world speech-language pathology settings.

The analysis code and the phone feature configurations for 12 languages are public at \url{https://github.com/Scott-Morgan-Foundation/phonological-subspace-severity} (tag \texttt{paper2-csl-revision-2026-11-r7}, directory \texttt{paper2-revision/}). The derived per-speaker inputs are available subject to the data-use agreements of the underlying corpora (see Data Availability).

\subsection{Declaration of Generative AI Use}

Bernard Muller (first author, Technical PI) communicates exclusively via an AI-assisted eye-gaze interface due to motor neurone disease. Generative AI (Claude, Anthropic) was used as an assistive communication tool throughout the research process: experimental design, code development, data analysis, statistical computation, figure generation, and manuscript drafting. All scientific decisions, interpretations, and conclusions were made by the authors. The AI served as an accessibility tool enabling a researcher with severe physical disability to conduct computational research---analogous to a screen reader for visually impaired researchers. No AI-generated content was presented without author review and verification.

\subsection{Ethics and Data Governance}

This study is a secondary analysis of previously collected speech materials accessed under the licences, permissions, or data-use conditions detailed below. No participants were recruited or interacted with for the present study. The YouTube-derived corpora, sourced from publicly released documentaries, are the one exception to the research-dataset framing and are described, with the basis on which they are used, at the end of this section.

The following corpora are publicly available under open Creative Commons licences: LibriSpeech [21] from OpenSLR under CC BY 4.0; SLR65 Tamil from OpenSLR under CC BY-SA 4.0; SVD (Saarbruecken Voice Database) publicly downloadable under CC BY 4.0; CDLI Kenyan Swahili under CC BY 4.0; Hungarian healthy controls from Mozilla Common Voice under CC0; Neurovoz [29] publicly available on Zenodo under CC BY 4.0; and IPVS publicly available on HuggingFace under CC BY 4.0.

The following corpora are publicly available under Creative Commons NonCommercial or NoDerivatives licences: MDSC [33] under CC BY-NC 4.0 via the SLT 2024 LRDWWS Challenge distribution channel referenced on the AISHELL Tech custodian page (\url{https://www.aishelltech.com/AISHELL_6B}); EasyCall under CC BY-NC 4.0; and AVFAD publicly available at talkbank.org/motor under CC BY-NC-ND 4.0, the ND term restricting derivative-clip release. VOC-ALS [45] was accessed through its official Synapse repository (project \texttt{syn53009474}) under the custodian's applicable research-only data-use conditions, without a bespoke agreement with the authors; it is referenced here only for language-coverage completeness and is not part of the 3,374-speaker analysis pipeline. EWA-DB was accessed via Zenodo under the custodian's research-only EULA. TORGO [22] was obtained by public download from the corpus distribution site under its stated terms of use; UA-Speech [23] under written permission from Prof.\ Mark Hasegawa-Johnson (University of Illinois), 6 February 2026; Domotica under email permission from Prof.\ Helmer Strik (Radboud University); CHASING under email permission from Radboud University (DOI: 10.1111/1460-6984.12722); and TreasureHunters1 under email permission from Radboud University (DOI: 10.35273/0w4g-bb09). The following corpora were obtained under signed data-use, data-sharing, or academic licences: SAP [46] from the University of Illinois under a signed data use agreement; COPAS [26] under a signed licence from the Dutch Language Institute (IVDNT); CDSD under a signed academic licence from Prof.\ Su-Jing Wang, \textcopyright\ Su-Jing Wang credit required, no redistribution; the Hungarian Dysarthria Database from the Budapest University of Technology and Economics under a signed research licence; SSNCE Tamil [47] through the Linguistic Data Consortium (LDC2021S04); and PC-GITA [30] under a signed research-use licence with the GITA laboratory at the Universidad de Antioquia, executed by co-author Prof.\ Antonio Ortiz Barra\~n\'on (Tec de Monterrey) on 4 March 2026.

The YouTube-derived corpora (YouTube\_French and YouTube\_German) comprise 24 French speaker codes (4 ALS codes from 3 individuals, 20 controls) and 13 German speaker codes (4 ALS codes from 3 individuals, 9 controls), sourced from publicly released documentaries (four per language). No individual research consent was obtained from the featured speakers. Speaker identifiers used in the analysis are coded, no mapping from those codes to identities is published, and no audio is redistributed. For copyright purposes, the automated extraction of acoustic features relied on the text-and-data-mining framework in Article 3 of Directive (EU) 2019/790, insofar as applicable to a qualifying research organisation with lawful access to the source material. Data-protection compliance is governed separately by the GDPR and applicable Member-State law; Article 89 GDPR provides safeguards for scientific research but does not itself establish the legal basis for processing.

Data governance. The CANDOR research programme operates under a patient-centred data governance framework aligned with GDPR principles. All datasets are accessed under the terms specified by their original custodians, and no raw audio is redistributed or stored beyond the scope of the original data sharing agreements. The analysis pipeline processes audio into de-identified aggregate statistical features (d$'$ scores and structural metrics); the outputs retain only each corpus's own pseudonymous speaker codes, not audio. The analysis code and phone feature configurations are public at \url{https://github.com/Scott-Morgan-Foundation/phonological-subspace-severity} (tag \texttt{paper2-csl-revision-2026-11-r7}, directory \texttt{paper2-revision/}). Reproducing the reported numbers also requires the derived per-speaker inputs, which remain under the data-use agreements of the underlying corpora, so that speaker-level data stays within each corpus's original consent and licensing framework.

\begingroup\small\sloppy
\subsection{Data Availability \& Reproducibility Statement}

The analysis code and the phone feature configurations for 12 languages are public at \url{https://github.com/Scott-Morgan-Foundation/phonological-subspace-severity} (tag \texttt{paper2-csl-revision-2026-11-r7}, directory \texttt{paper2-revision/}); the repository README maps each table and figure to the script that generates it and lists the configuration files of the current reproduction route; the frozen submission package is preserved unchanged under earlier tags. The Hungarian values are those of the submitted analysis; the configuration that produced them was not preserved. Re-extraction with the released Hungarian configuration agrees closely with them for the 28 Hungarian healthy controls (Spearman 0.983--1.000) but only approximately for the 31 speakers of the Hungarian dysarthric corpus (Spearman 0.851, median absolute difference 0.19); the cause of this difference is unconfirmed. Three of those speakers (one PD, two stroke) enter the six-group analyses, and none enters the fixed-token analysis. We re-ran every analysis behind the reported numbers with the re-extracted Hungarian values, without the Hungarian dysarthric corpus, and without all 59 Hungarian speakers: no reported conclusion changes. With the re-extracted values, changes are confined to the last printed digit (for example, macro F1 21.6\% rather than 21.9\%), except that the same-language profile-shape test gives p = 0.042 rather than 0.028. The analysis scripts, and a file that maps the numbers in the abstract, contributions, tables, figure captions and conclusion to the saved outputs that produce them (other numbers are traced by value), are included in the verification package available from the corresponding author. Access to the underlying corpora is governed by the licences and agreements listed in the Ethics and Data Governance section; several require signed data-use agreements that do not permit redistribution. The derived speaker-level tables (the master results file and the five backbone files) are therefore not public; they are available from the corresponding author to researchers who hold access to the underlying corpora, subject to those agreements.

\paragraph{Software and model identifiers.} Forced alignment used Montreal Forced Aligner v3.1.3 (Kaldi backend) with language-specific acoustic models from the MFA repository: \texttt{english\_mfa} v3.1.0, \texttt{dutch\_cv} v2.1.0, \texttt{spanish\_mfa} v3.1.0, \texttt{french\_mfa} v3.1.0, \texttt{mandarin\_mfa} v3.1.0, \texttt{italian\_cv} v2.1.0, \texttt{czech\_mfa} v3.1.0 (used for Slovak), \texttt{hungarian\_cv} v2.1.0, \texttt{portuguese\_mfa} v3.1.0, \texttt{german\_mfa} v3.1.0, \texttt{swahili\_mfa} v3.1.0, and \texttt{tamil\_cv} v2.1.0. Tamil was additionally aligned with torchaudio MMS forced alignment ([48]; \texttt{facebook/mms-1b-all}) where MFA romanisation failed.

SSL backbones were \texttt{facebook/hubert-base-ls960} (HuBERT-base, 94.4M params), \texttt{facebook/hubert-large-ls960-ft*} (HuBERT-large, 315.4M params; fine-tuned checkpoint used frozen), \texttt{microsoft/wavlm-base} (WavLM-base, 94.4M params), \texttt{facebook/wav2vec2-base} (wav2vec2-base, 94.4M params), \texttt{facebook/wav2vec2-xls-r-300m} (XLS-R, 315.4M params), and \texttt{facebook/mms-300m} (MMS, 315.4M params). All models were used frozen; embeddings were extracted from the final hidden layer at 50 Hz (20 ms frame rate). CTC confidence models were \texttt{jonatasgrosman/wav2vec2-large-xlsr-53-\{language\}} for each of the 12 languages [43].

\paragraph{Preprocessing and outputs.} All audio was resampled to 16 kHz mono 16-bit PCM. The minimum phone-token threshold for valid d-prime estimation was 5 tokens per class. Phone-to-feature mappings were stored in language-specific JSON configuration files in the repository. The master speaker-level results CSV contains speaker identifier, dataset, language, aetiology, severity label and numeric code, control flag, token count, the 9 d-prime features, the 3 structural and 3 prosodic features, five voice-quality measures and the CTC-Conf score; a missing value marks a feature that could not be estimated.

\endgroup

\subsection{CRediT Author Contributions}

Bernard Muller: Conceptualization, Methodology, Software, Formal analysis, Investigation, Data curation, Writing---original draft, Visualization.

Antonio Armando Ortiz Barrañón: Validation, Software, Writing---review \& editing.

LaVonne Roberts: Supervision, Funding acquisition, Project administration, Writing---review \& editing.

\section*{References}\addcontentsline{toc}{section}{References}
{\small\sloppy
\renewcommand{\labelenumi}{[\arabic{enumi}]}
\begin{enumerate}
\item Duffy, J. R. (2019). Motor Speech Disorders: Substrates, Differential Diagnosis, and Management (4th ed.). Elsevier.

\item Muller, B., Ortiz Barranon, A. A., and Roberts, L. (2026). Training-free cross-lingual dysarthria severity assessment via phonological subspace analysis in self-supervised speech representations. arXiv preprint arXiv:2604.10123. \href{https://doi.org/10.48550/arXiv.2604.10123}{\nolinkurl{doi:10.48550/arXiv.2604.10123}}

\item Choi, K., Yeo, E., Cho, C. J., Mortensen, D. R., and Harwath, D. (2026). Self-supervised speech models encode phonetic context via position-dependent orthogonal subspaces. arXiv preprint arXiv:2603.12642.

\item Cho, C. J., Wu, P., Mohamed, A., and Anumanchipalli, G. K. (2023). Evidence of vocal tract articulation in self-supervised learning of speech. In Proceedings of ICASSP 2023, 1--5. \href{https://doi.org/10.1109/ICASSP49357.2023.10094711}{\nolinkurl{doi:10.1109/ICASSP49357.2023.10094711}}

\item Halpern, B. M., Tienkamp, T., Abur, D., and Toda, T. (2026). PathBench: Speech intelligibility benchmark for automatic pathological speech assessment. arXiv preprint arXiv:2603.08097. \href{https://doi.org/10.48550/arXiv.2603.08097}{\nolinkurl{doi:10.48550/arXiv.2603.08097}}

\item Kadirvelu, B., Stumpf, L., Waibel, S., and Faisal, A. A. (2025). Speaker-independent dysarthria severity classification using self-supervised transformers and multi-task learning. PLOS Digital Health, 4(11), e0001076. \href{https://doi.org/10.1371/journal.pdig.0001076}{\nolinkurl{doi:10.1371/journal.pdig.0001076}}

\item Joshy, A. A. and Rajan, R. (2022). Automated dysarthria severity classification: A study on acoustic features and deep learning techniques. IEEE Transactions on Neural Systems and Rehabilitation Engineering, 30, 1147-1157. https://doi.org/10.1109/TNSRE.2022.3169814
\item Hernandez, A., Yeo, E., Choi, K., Li, C.-J., Yue, Z., Das, R. K., Rusz, J., Magimai Doss, M., Orozco-Arroyave, J. R., Arias-Vergara, T., Maier, A., Noth, E., Mortensen, D. R., Harwath, D., and Perez-Toro, P. A. (2026). Adapting self-supervised speech representations for cross-lingual dysarthria detection in Parkinson's disease. arXiv:2603.22225.

\item Rios-Urrego, C. D., Rusz, J., and Orozco-Arroyave, J. R. (2024). Automatic speech-based assessment to discriminate Parkinson's disease from essential tremor with a cross-language approach. npj Digital Medicine, 7, 37. \href{https://doi.org/10.1038/s41746-024-01027-6}{\nolinkurl{doi:10.1038/s41746-024-01027-6}}

\item Yeo, E. J., Liss, J. M., Berisha, V., and Mortensen, D. R. (2026). Multilingual dysarthric speech assessment using universal phone recognition and language-specific phonemic contrast modeling. arXiv preprint arXiv:2601.21205. \href{https://doi.org/10.48550/arXiv.2601.21205}{\nolinkurl{doi:10.48550/arXiv.2601.21205}}

\item Baevski, A., Zhou, H., Mohamed, A., and Auli, M. (2020). wav2vec 2.0: A framework for self-supervised learning of speech representations. In Advances in Neural Information Processing Systems, 33, 12449--12460.

\item Babu, A., Wang, C., Tjandra, A., Lakhotia, K., Xu, Q., Goyal, N., Singh, K., von Platen, P., Saraf, Y., Pino, J., Baevski, A., Conneau, A., and Auli, M. (2022). XLS-R: Self-supervised cross-lingual speech representation learning at scale. In Proceedings of Interspeech 2022, 2278--2282. \href{https://doi.org/10.21437/Interspeech.2022-143}{\nolinkurl{doi:10.21437/Interspeech.2022-143}}

\item Violeta, L. P., Huang, W.-C., and Toda, T. (2022). Investigating self-supervised pretraining frameworks for pathological speech recognition. In Proceedings of Interspeech 2022. \href{https://doi.org/10.21437/Interspeech.2022-10043}{\nolinkurl{doi:10.21437/Interspeech.2022-10043}}

\item Sapkota, B., Shrestha, S., and Baral, R. (2025). Do all features matter? Layer-wise feature probing of self-supervised speech models for dysarthria severity classification. Speech Communication, 175, 103326. \href{https://doi.org/10.1016/j.specom.2025.103326}{\nolinkurl{doi:10.1016/j.specom.2025.103326}}

\item Bae, J., Zheng, X., Kim, M., Yoo, C. D., and Hasegawa-Johnson, M. (2026). Something from nothing: Data augmentation for robust severity level estimation of dysarthric speech. arXiv:2603.15988.

\item Javanmardi, F., Arias-Vergara, T., Orozco-Arroyave, J. R., and Nöth, E. (2024). Pre-trained models for detection and severity level classification of dysarthria from speech. Speech Communication, 156, 103047. \href{https://doi.org/10.1016/j.specom.2024.103047}{\nolinkurl{doi:10.1016/j.specom.2024.103047}}

\item McAuliffe, M., Socolof, M., Mihuc, S., Wagner, M., and Sonderegger, M. (2017). Montreal Forced Aligner: Trainable text-speech alignment using Kaldi. In Proceedings of Interspeech 2017, 498--502. \href{https://doi.org/10.21437/Interspeech.2017-1386}{\nolinkurl{doi:10.21437/Interspeech.2017-1386}}

\item Hsu, W.-N., Bolte, B., Tsai, Y.-H. H., Lakhotia, K., Salakhutdinov, R., and Mohamed, A. (2021). HuBERT: Self-supervised speech representation learning by masked prediction of hidden units. IEEE/ACM Transactions on Audio, Speech, and Language Processing, 29, 3451--3460. \href{https://doi.org/10.1109/TASLP.2021.3122291}{\nolinkurl{doi:10.1109/TASLP.2021.3122291}}

\item Macmillan, N. A. and Creelman, C. D. (2005). Detection Theory: A User's Guide (2nd ed.). Lawrence Erlbaum Associates.

\item Hasegawa-Johnson, M., Zheng, X., Kim, H., Mendes, C., Dickinson, M., Hege, E., Zwilling, C., Moore Channell, M., Mattie, L., Hodges, H., Ramig, L., Bellard, M., Shebanek, M., Sari, L., Kalgaonkar, K., Frerichs, D., Bigham, J. P., Findlater, L., Lea, C., Herrlinger, S., Korn, P., Abou-Zahra, S., Heywood, R., Tomanek, K., and MacDonald, B. (2024). Community-supported shared infrastructure in support of speech accessibility. Journal of Speech, Language, and Hearing Research, 67(11), 4162--4175. \href{https://doi.org/10.1044/2024_JSLHR-24-00122}{\href{https://doi.org/10.1044/2024}{\nolinkurl{doi:10.1044/2024}}\_JSLHR-24-00122}

\item Panayotov, V., Chen, G., Povey, D., and Khudanpur, S. (2015). LibriSpeech: An ASR corpus based on public domain audio books. In Proceedings of ICASSP 2015, 5206--5210. \href{https://doi.org/10.1109/ICASSP.2015.7178964}{\nolinkurl{doi:10.1109/ICASSP.2015.7178964}}

\item Rudzicz, F., Namasivayam, A. K., and Wolff, T. (2012). The TORGO database of acoustic and articulatory speech from speakers with dysarthria. Language Resources and Evaluation, 46(4), 523--541. \href{https://doi.org/10.1007/s10579-011-9145-0}{\nolinkurl{doi:10.1007/s10579-011-9145-0}}

\item Kim, H., Hasegawa-Johnson, M., Perlman, A., Gunderson, J., Huang, T., Watkin, K., and Frame, S. (2008). Dysarthric speech database for universal access research. In Proceedings of Interspeech 2008, 1741--1744. \href{https://doi.org/10.21437/Interspeech.2008-480}{\nolinkurl{doi:10.21437/Interspeech.2008-480}}

\item Rusko, M., Sabo, R., Trnka, M., Zimmermann, A., Malaschitz, R., Ruzicky, E., Brandoburova, P., Kevicka, V., and Skorvanek, M. (2024). Slovak database of speech affected by neurodegenerative diseases. Scientific Data, 11, 1320. \href{https://doi.org/10.1038/s41597-024-04171-6}{\nolinkurl{doi:10.1038/s41597-024-04171-6}}

\item Jesus, L. M. T., Belo, I., Machado, J., and Hall, A. (2017). The advanced voice function assessment databases (AVFAD): Tools for voice clinicians and speech research. In Advances in Speech-language Pathology. IntechOpen. \href{https://doi.org/10.5772/intechopen.69643}{\nolinkurl{doi:10.5772/intechopen.69643}}

\item Middag, C., Martens, J.-P., Van Nuffelen, G., and De Bodt, M. (2009). Automated intelligibility assessment of pathological speech using phonological features. EURASIP Journal on Advances in Signal Processing, 2009, 1--9. \href{https://doi.org/10.1155/2009/629030}{\nolinkurl{doi:10.1155/2009/629030}}

\item Ganzeboom, M., Bakker, M., Beijer, L., Strik, H., and Rietveld, T. (2022). A serious game for speech training in dysarthric speakers with Parkinson's disease: Exploring therapeutic efficacy and patient satisfaction. International Journal of Language and Communication Disorders, 57(5), 1091--1106. \href{https://doi.org/10.1111/1460-6984.12722}{\nolinkurl{doi:10.1111/1460-6984.12722}}

\item Ganzeboom, M., Bakker, M., Beijer, L., Rietveld, T., and Strik, H. (2018). Speech training for neurological patients using a serious game. British Journal of Educational Technology, 49(4), 761--774. \href{https://doi.org/10.1111/bjet.12640}{\nolinkurl{doi:10.1111/bjet.12640}}

\item Mendes-Laureano, J., Gomez-Garcia, J. A., Guerrero-Lopez, A., Luque-Buzo, E., Arias-Londono, J. D., Grandas-Perez, F. J., and Godino-Llorente, J. I. (2024). NeuroVoz: A Castilian Spanish corpus of parkinsonian speech. Scientific Data, 11, 1367. \href{https://doi.org/10.1038/s41597-024-04186-z}{\nolinkurl{doi:10.1038/s41597-024-04186-z}}

\item Orozco-Arroyave, J. R., Arias-Londono, J. D., Vargas-Bonilla, J. F., Gonzalez-Rativa, M. C., and Noth, E. (2014). New Spanish speech corpus database for the analysis of people suffering from Parkinson's disease. In Proceedings of LREC 2014, 342--347.

\item Dimauro, G., Di Nicola, V., Bevilacqua, V., Caivano, D., and Girardi, F. (2017). Assessment of speech intelligibility in Parkinson's disease using a speech-to-text system. IEEE Access, 5, 22199-22208. \href{https://doi.org/10.1109/ACCESS.2017.2762475}{\nolinkurl{doi:10.1109/ACCESS.2017.2762475}}

\item Turrisi, R., Braccia, A., Emanuele, M., Giulietti, S., Pugliatti, M., Sensi, M., Fadiga, L., and Badino, L. (2021). EasyCall corpus: A dysarthric speech dataset. In Proceedings of Interspeech 2021, 41--45. \href{https://doi.org/10.21437/Interspeech.2021-549}{\nolinkurl{doi:10.21437/Interspeech.2021-549}}

\item Gao, M., Chen, H., Du, J., Xu, X., Guo, H., Bu, H., Yang, J., Li, M., and Lee, C.-H. (2024). Enhancing voice wake-up for dysarthria: Mandarin Dysarthria Speech Corpus release and customized system design. In Proceedings of Interspeech 2024. \href{https://doi.org/10.21437/Interspeech.2024-879}{\nolinkurl{doi:10.21437/Interspeech.2024-879}}

\item Wan, Y., Sun, M., Kang, X., Li, J., Guo, P., Gao, M., and Wang, S.-J. (2024). CDSD: Chinese dysarthria speech database. In Proceedings of Interspeech 2024, 4109--4113. \href{https://doi.org/10.21437/Interspeech.2024-1597}{\nolinkurl{doi:10.21437/Interspeech.2024-1597}}

\item OpenSLR (2020). SLR65: Crowdsourced high-quality Tamil multi-speaker speech dataset {[}Dataset{]}. \url{https://www.openslr.org/65/} (accessed 2026-03-01)

\item Puetzer, M. and Barry, W. J. (2007). Saarbruecken Voice Database. Institute of Phonetics, Saarland University. \url{http://www.stimmdatenbank.coli.uni-saarland.de/}

\item Mihajlik, P., Toth, L., and Nemeth, G. (2023). Hungarian dysarthric speech database {[}Dataset{]}. Budapest University of Technology and Economics.

\item CDLI (2024). Kenyan Swahili Dysarthric Speech Corpus {[}Dataset{]}. Centre for Digital Language Inclusion, University of Cape Town. \url{https://www.cdli.uct.ac.za/} (accessed 2026-03-15)

\item Stipancic, K. L., Palmer, K. M., Rowe, H. P., Yunusova, Y., Berry, J. D., and Green, J. R. (2021). You say severe, I say mild: Toward an empirical classification of dysarthria severity. Journal of Speech, Language, and Hearing Research, 64(12), 4718--4735. \href{https://doi.org/10.1044/2021_JSLHR-21-00197}{\nolinkurl{doi:10.1044/2021\_JSLHR-21-00197}}

\item Chen, S., Wang, C., Chen, Z., Wu, Y., Liu, S., Chen, Z., Li, J., Kanda, N., Yoshioka, T., Xiao, X., Wu, J., Zhou, L., Ren, S., Qian, Y., Qian, Y., Wu, J., Zeng, M., Yu, X., and Wei, F. (2022). WavLM: Large-scale self-supervised pre-training for full stack speech processing. IEEE Journal of Selected Topics in Signal Processing, 16(6), 1505--1518. \href{https://doi.org/10.1109/JSTSP.2022.3188113}{\nolinkurl{doi:10.1109/JSTSP.2022.3188113}}

\item Westfall, J. and Yarkoni, T. (2016). Statistically controlling for confounding constructs is harder than you think. PLOS ONE, 11(3), e0152719. \href{https://doi.org/10.1371/journal.pone.0152719}{\nolinkurl{doi:10.1371/journal.pone.0152719}}

\item Cohen, J. (1988). Statistical Power Analysis for the Behavioral Sciences (2nd ed.). Lawrence Erlbaum Associates.

\item Grosman, J. (2021). Fine-tuned XLSR-53 large models for speech recognition {[}Model collection{]}. HuggingFace. \url{https://huggingface.co/jonatasgrosman} (accessed 2026-04-01)

\item Van Nuffelen, G., Middag, C., De Bodt, M., and Martens, J.-P. (2009). Speech technology-based assessment of phoneme intelligibility in dysarthria. International Journal of Language and Communication Disorders, 44(5), 716--730. \href{https://doi.org/10.1080/13682820802342062}{\nolinkurl{doi:10.1080/13682820802342062}}

\item Dubbioso, R., Spisto, M., Verde, L., Iuzzolino, V. V., Senerchia, G., Salvatore, E., De Pietro, G., De Falco, I., and Sannino, G. (2024). Voice signals database of ALS patients with different dysarthria severity and healthy controls. Scientific Data, 11(1), 800. \href{https://doi.org/10.1038/s41597-024-03597-2}{\nolinkurl{doi:10.1038/s41597-024-03597-2}}

\item Zheng, X., Phukon, B., Na, J., Cutrell, E., Han, K. J., Hasegawa-Johnson, M., Jiang, P.-P., Kuila, A., Lea, C., MacDonald, B., Mantena, G., Ravichandran, V., Sari, L., Tomanek, K., Yoo, C. D., and Zwilling, C. (2025). The Interspeech 2025 Speech Accessibility Project Challenge. In Proceedings of Interspeech 2025, 3269--3273. \href{https://doi.org/10.21437/Interspeech.2025-566}{\nolinkurl{doi:10.21437/Interspeech.2025-566}}

\item LDC (2021). The SSNCE Database of Tamil Dysarthric Speech {[}Dataset{]}. Linguistic Data Consortium, LDC2021S04. \href{https://doi.org/10.35111/hkh2-vh40}{\nolinkurl{doi:10.35111/hkh2-vh40}}

\item Pratap, V., Tjandra, A., Shi, B., Tomasello, P., Babu, A., Kundu, S., Elkahky, A., Ni, Z., Vyas, A., Fazel-Zarandi, M., Baevski, A., Adi, Y., Zhang, X., Hsu, W.-N., Conneau, A., and Auli, M. (2024). Scaling speech technology to 1,000+ languages. Journal of Machine Learning Research, 25(97), 1--52.

\end{enumerate}
}

\end{document}